\documentclass{article}

\newif\ifappendix
\appendixfalse
\ifappendix
\else

\usepackage{xr}
\makeatletter

\newcommand*{\addFileDependency}[1]{
\typeout{(#1)}
\@addtofilelist{#1}
\IfFileExists{#1}{}{\typeout{No file #1.}}
}\makeatother

\newcommand*{\myexternaldocument}[1]{%
\externaldocument{#1}%
\addFileDependency{#1.tex}%
\addFileDependency{#1.aux}%
}
    \myexternaldocument{appendix}
\fi

\usepackage[final]{corl_2022}
\usepackage{times}
\usepackage{subcaption}

\usepackage[numbers]{natbib}
\usepackage{multicol}

\usepackage{graphics} 
\usepackage{epsfig} 
\usepackage{times} 
\usepackage{amsmath} 
\usepackage{amssymb}  
\usepackage[dvipsnames]{xcolor}
\usepackage[ruled,vlined,linesnumbered]{algorithm2e}
\usepackage{multirow}
\usepackage{xr-hyper}

\usepackage{blindtext}
\usepackage{etoc} 
\usepackage{bibunits}

\renewcommand{\contentsname}{Appendix}

\usepackage{color, soul} 

\usepackage[font={footnotesize}]{caption}
\usepackage{hyperref}
\usepackage{wrapfig}
\usepackage{xspace}
\usepackage{floatrow}
\newfloatcommand{capbtabbox}{table}[][\FBwidth]

\begin{document}

\title{Deep Projective Rotation Estimation 
through Relative Supervision}

\author{
  Brian Okorn\thanks{Equal Contribution}, Chuer Pan\footnotemark[1],
  Martial Hebert,
  David Held \\
  Robotics Institute, School of Computer Science\\
  Carnegie Mellon University, 
  United States\\
  \texttt{\{bokorn, chuerp, mhebert, dheld\}@andrew.cmu.edu} \\
}



%

\maketitle

\begin{abstract}
Orientation estimation is the core to a variety of vision and robotics tasks such as camera and object pose estimation. Deep learning has offered a way to develop image-based orientation estimators; however, such estimators often require training on a large labeled dataset, which can be time-intensive to collect. In this work, we explore whether self-supervised learning from unlabeled data can be used to alleviate this issue.  Specifically, we assume access to estimates of the relative orientation between neighboring poses, such that can be obtained via a local alignment method. While self-supervised learning has been used successfully for translational object keypoints, in this work, we show that naively applying relative supervision to the rotational group $SO(3)$ will often fail to converge due to the non-convexity of the rotational space. To tackle this challenge, we propose a new algorithm for self-supervised orientation estimation which utilizes Modified Rodrigues Parameters to stereographically project the closed manifold of $SO(3)$ to the open manifold of $\mathbb{R}^{3}$, allowing the optimization to be done in an open Euclidean space. We empirically validate the benefits of the proposed algorithm for rotational averaging problem in two settings: (1) direct optimization on rotation parameters, and (2) optimization of parameters of a convolutional neural network that predicts object orientations from images. In both settings, we demonstrate that our proposed algorithm is able to converge to a consistent relative orientation frame much faster than algorithms that purely operate in the $SO(3)$ space. Additional information can be found on our \href{https://sites.google.com/view/deep-projective-rotation/home}{website}. 
\end{abstract}



\section{Introduction}
Pose estimation is a critical component for a wide variety of computer vision and robotic tasks. It is a common primitive for grasping, manipulation, and planning tasks. For motion planning and control, estimating an object's pose can help a robot avoid collisions or plan how to use the object for a given task. 
The current top performing methods for pose estimation use machine learning to estimate the object's pose from an image; however, training these estimators tends to rely on direct supervision of the object orientation~\cite{xiang2018posecnn,labbe2020cosypose,wang2019densefusion}. Obtaining such supervision can be difficult and requires either time-consuming annotations or synthetic data, which might differ from the real world.
In this work, we explore whether self-supervised learning can be used to alleviate this issue by training an object orientation estimator from unlabeled data. Specifically, we assume that we can estimate the relative rotation of an object between neighboring object poses in a self-supervised manner. Such relative supervision can be easily obtained in practice, for example through a local registration method such as Iterative Closest Point (ICP)~\cite{zhou2016fast} or camera pose estimation. 


Relative self-supervision has been previously used for representation learning in estimating translational keypoints~\cite{manuelli2019kpam, sun2018integral, suwajanakorn2018discovery}. These methods use only relative supervision to ensure that the keypoints are consistent across views of the object, and do not directly supervise the keypoint locations. 
In this work, we explore whether such relative self-supervision can similarly be used in estimating object orientations.
We show that naively applying such relative supervision to rotations on the $SO(3)$ manifold will often fail to converge. Unlike self-supervised learning of translational keypoints, the rotational averaging problem~\cite{hartley2013rotation} is inherently non-convex, with many local optima. While there exist global optimization algorithms which jointly optimize all pairs of rotations for this problem~\cite{dellaert2020shonan,wang2013exact}, they are not easily integrated into the iterative, stochastic gradient descent methods used to train neural network-based pose estimators.

To address this issue, we propose a new algorithm, Iterative Modified Rodrigues Projective Averaging,
which uses Modified Rodrigues Parameters to map from the closed manifold of $SO(3)$ to the open space of $\mathbb{R}^3$.
In doing so, we obtain faster convergence with a lower likelihood of falling into local optima. 
Our experiments show that our method converges faster and more consistently than the standard $SO(3)$ optimization and can easily be integrated into a neural network training pipeline. Additionally, in the Appendix~\ref{appendix:theory}, we include an intuitive theoretical example describing how, while not all local optima are removed, the dimensionality of a set of problematic configurations is greatly reduced when optimizing using our algorithm, as compared to optimizing in the space of $SO(3)$.

The primary contributions of this work are:
\begin{itemize}
    \item We propose a new algorithm, Iterative Modified Rodrigues Projective Averaging, which is an iterative 
    method for learning rotation estimation using only relative supervision and can be applied to neural network optimization.
    \item We  empirically investigate the convergence behavior of our algorithm as compared to optimizing on the $SO(3)$ manifold.
    \item We demonstrate that our algorithm can be used to train a neural network-based pose estimator using only relative supervision. 
\end{itemize}
\section{Related Work}

\textbf{Averaging and Consensus Estimation:}
Consensus methods, sometimes referred to as averaging methods, have a long history of research. The goal of these methods is, given a distributed set of estimates, to produce a consistent prediction of a value using relative information. While there are iterative algorithms with good convergence properties in Euclidean space~\cite{degroot1974reaching,chatterjee1977towards,olshevsky2009convergence,russell2011optimal,beck2015weiszfeld}, optimizing over the closed manifold of $SO(3)$ can be more difficult, as the region is non-convex, with many local minima. 
~\citet{hartley2013rotation, hartley2011l1} describe several methods of finding a consistent set of rotations, though their convergence is similarly not guaranteed outside of a radius $r \leq \frac{\pi}{2}$ ball in $SO(3)$. 
\citet{wang2013exact} find an exact solution to this problem, using a combination of a semidefinite programming relaxation and a robust penalty function. More recently, Shonan Rotation Averaging~\cite{dellaert2020shonan} shows that projecting to higher dimensional spaces allows for the recovery of a globally optimal solution using semidefinite programming. \citet{chatterjee2013efficient, chatterjee2017robust} use iterative re-weighted least-squares to recover a global optimal solution using global error estimates. \citet{shi2020message} extends this work, using cycle consistency and message passing.  
These solutions require global error estimates or semidefinate programming, which are incompatible with the stochastic gradient descent methods used to train neural networks.

\textbf{Supervised Orientation Estimation:}
Past work has explored using a neural network to predict an object's orientation. 
Traditionally, these methods rely on supervising the rotations using a known absolute orientation, whether in the form of quaternions~\cite{kendall2015posenet,xiang2018posecnn, kendall2017geometric}, axis-angle~\cite{do2018deep}, or Euler angles~\cite{su2015render}. More recently, 6D~\cite{zhou2019continuity, labbe2020cosypose}, 9D~\cite{levinson2020analysis}, and 10D~\cite{peretroukhin2020smooth} representations have been developed for continuity and smoothness. Recently,~\citet{terzakis2018modified} introduced Modified Rodrigues Parameters, a projection of the unit quaternion sphere $\mathbb{S}^3$ to $\mathbb{R}^3$ used in attitude control~\cite{crassidis1996attitude}, to a range of common computer vision problems.
~\citet{terzakis2018modified} does not, however, address the unique problems found in the rotation averaging problem.


Recently, there has been research into mapping the Riemannian optimization to the Euclidean optimization used for network training~\cite{chen2021projective,teed2021tangent, bregier2021deep, arora2009learning, bonnabel2013stochastic}. These methods focus on applying tangent space gradients from losses in 3D transformation groups. Specifically, Projective Manifold Gradient Layer~\cite{chen2021projective} ensures that the gradients take into account any projection operations, such that the gradients point towards the nearest valid representation in the projection's preimage. While this does map the Riemannian optimization into a Euclidean problem, it does not solve the problems caused by the closed manifold of $SO(3)$, as this does not alter the underlying topology of this manifold.
\section{Problem Definition}
\label{sec:problem_def}
We formally describe the problem of self-supervised 
orientation estimation below.
We assume that we are given a set of inputs observations $\{I_1, \dots, I_N\}$, of an object where, in each input observation $I_i$, the object is viewed from an unknown orientation $R_i$. These inputs could be in the form of images, point clouds, or some other object representation. 
While we do not know the absolute object orientations $R_i$ in any reference frame, we assume that we do know a subset of the relative rotations $R_j^i$, possibly from a local registration method like ICP, between the object in images $I_j$ and $I_i$, such that $R_i = R_i^j R_j$. 
Our goal is to learn a function $f(I_i)$ that estimates an orientation of the object in each image, $f(I_i) = \hat{R}_i$ that minimizes the pairwise error between all input pairs and their ground truth relative rotations, with respect to the geodesic distance metric $d(R_i, R_j) = \| \log(R_i^\top R_j) \|^2$.  Given a  set of rotations $\mathcal{R}=\{R_1, \dots, R_N\}$, the core optimization objective is thus: 
\begin{equation}
    \min_{\hat{R}_i,\hat{R}_j \in \mathcal{R}} \sum_{i,j} d( \hat{R_i}, R_i^j \hat{R_j})
\label{eq:distance_minimization}
\end{equation}
Note that this optimization does not have a unique solution, since the solution $\hat{R}_i:=S R_i, \forall i$ minimizes this error for any constant rotation $S$. 
\looseness=-1
In many robotics tasks, relative rotations can be accurately estimated only when their magnitude is small as many registration algorithms, such as ICP, requires a good initialization near the optimum. Following this observation, we assume that we can only accurately supervise relative rotations when they are small in magnitude. 
This leads to a local neighborhood structure where each rotation $R_i$ is connected to $R_j$ only in a local neighborhood around $R_i$, when $d(R_i, R_j) < \epsilon$, and the set of all $R_j$'s connected to $R_i$ form the neighborhood set of $\mathcal{N}_i$. 
While the algorithms described in this manuscript do not rely on this angle $\epsilon$, it can be scaled as needed based on the accuracy of the relative rotation estimation method (e.g. ICP, etc).

Our eventual goal is to represent the function $f(I_i)= \hat{R}_i$ as a neural network. Thus, we restrict the methods with which we compare to iterative methods that are updated using only a sampled subset of the rotations (as opposed to methods that perform a global optimization over the entire set of rotations $\{R_1, \dots, R_N\}$). This requirement is to match the conditions required by stochastic gradient descent, the primary method of training neural networks.


\section{Baselines}

\textbf{Preliminaries.}
\label{sec:preliminaries}
The 3D rotational space of $SO(3)\triangleq \{R\in \mathbb{R}^{3\times 3}: R^{\top}R = \mathbb{I}_{3\times 3}, \det{\left(R\right)} = 1\}$ is a compact matrix Lie group, which topologically is a compact manifold. 
Due to the compactness of the $SO(3)$ manifold, there exist configurations of pairs of points where multiple, non-unique geodesically minimal paths exist between them; for instance, there are two unique geodesically minimal paths for a pair of antipodal points on a circle, and there are infinitely many for a pair of antipodal points on a sphere.  This is not the case in an open manifold like the 3D Euclidean space of $\mathbb{R}^{3}$, over which there exists a unique geodesically minimal path between any arbitrary pair of points. The distinction in compactness between the 3D rotational space of $SO(3)$ and 3D Euclidean space makes optimization over $SO(3)$ more ill-conditioned than over the space of $\mathbb{R}^{3}$. This results in the optimization over the rotational space being non-convex. 
These properties of the $SO(3)$ manifold will affect the convergence of self-supervised orientation estimation, which we discuss below.

While self-supervised learning for objects translation, specifically in the form of object keypoints~\cite{manuelli2019kpam, sun2018integral, suwajanakorn2018discovery}, has shown great success, in this work, we show that naively applying such an iterative self-supervised formulation to the rotational group $SO(3)$ will often fail to converge. 
Below we discuss two approaches to self-supervised orientation estimation in $SO(3)$. 

\textbf{Quaternion Averaging:}
\label{sec:quat_consensus}
A standard objective in rotation estimation is to minimize the geodesic distance between a predicted unit quaternion and its corresponding ground-truth orientation~\cite{mahendran20173d, hartley2013rotation}, $\theta =  \arccos(2\langle \hat{q}_{i}, q_{gt} \rangle^2)$
%
where $\hat{q}_{i}$ is the predicted orientation for image $i$ and $q_{gt}$ is the ground-truth orientation. An objective function is often defined to directly minimize this geodesic distance~\cite{mahendran20173d}. 



\looseness=-1
In our task, defined above (Section~\ref{sec:problem_def}), we are given the relative rotation $q_i^j$ between some pairs of rotations $q_i$ and $q_j$. Using this relative supervision, we can use the geodesic distance between a sampled estimate, $\hat{q}_i$, its desired relative position 
with respect to a sampled neighbor and a known relative rotation $q_i^j$, $\tilde{q}_i = q_i^j \otimes \hat{q}_j$, leading to the loss  
\begin{equation}
    \mathcal{L}_q = 1-\langle \hat{q}_i, q_i^j \otimes \hat{q}_j \rangle^2
    \label{eq:quat_loss}
\end{equation}
where $\otimes$ denotes the quaternion multiplication.
Note that this loss  is monotonically related to the geodesic  distance 
when using unit quaternions, while avoiding the need to compute an $\arccos$. 

\textbf{$\mathbf{SO(3)}$ Averaging:}
\label{sec:so3_consensus}
To optimize the rotations with respect to the non-Euclidean geometry of the rotational manifold of $SO(3)$, one approach is 
described by \citet{manton2004globally}. Each orientation is iteratively updated in the tangent 
space using the logmap of $SO(3)$ and projected back to $SO(3)$ using the exponential map. 
Specifically, we can take the gradient of the loss 
\begin{subequations}
\begin{minipage}{.5\textwidth}
    \begin{equation} 
        \mathcal{L}_{SO(3)} = \left\|\log \left( R_i^\top R_i^j R_j\right)\right\|^2
    \label{eq:so3_loss}
    \end{equation}
\end{minipage}%
\begin{minipage}{.5\textwidth}
    \begin{equation} 
        \nabla_{\hat{r}_i} \mathcal{L}_{SO(3)} = r_\Delta = \log \left( R_i^\top R_i^j R_j\right)
    \end{equation}
\end{minipage}

\end{subequations}

which gives the update step $\hat{R}_i \gets \hat{R}_i \exp(\gamma r_\Delta)$, where $\gamma$ is the learning rate and $\log$ is the logmap of $SO(3)$. 
When optimizing the full set of orientations, this algorithm can fall into local optima due to the closed nature of the space which allows any orientation to be reached by more than one unique straight paths, as the space wraps around on itself.

\section{Method}
\looseness=-1
We propose an alternative that projects the optimization to an open image and optimizes the distances in that space. Specifically, we use the Modified Rodriguez Projection to minimize the relative error between neighboring poses in $\mathbb{R}^3$. 
We provide experiments in Section~\ref{sec:experiments} that show that self-supervised orientation estimation using Modified  Rodriguez Projection converges much faster than self-supervised orientation estimation in $SO(3)$, with theoretic analysis of an illustrative example available in the Appendix~\ref{appendix:theory}.

\subsection*{Iterative Modified Rodrigues Projective Averaging}
\label{sec:mrp_consensus}
\begin{wrapfigure}[16]{r}{0.5\textwidth}
 	\centering
 	 	\vspace{-1em}
    \includegraphics[width=1.0 \textwidth,trim={70 30 10 80},clip]{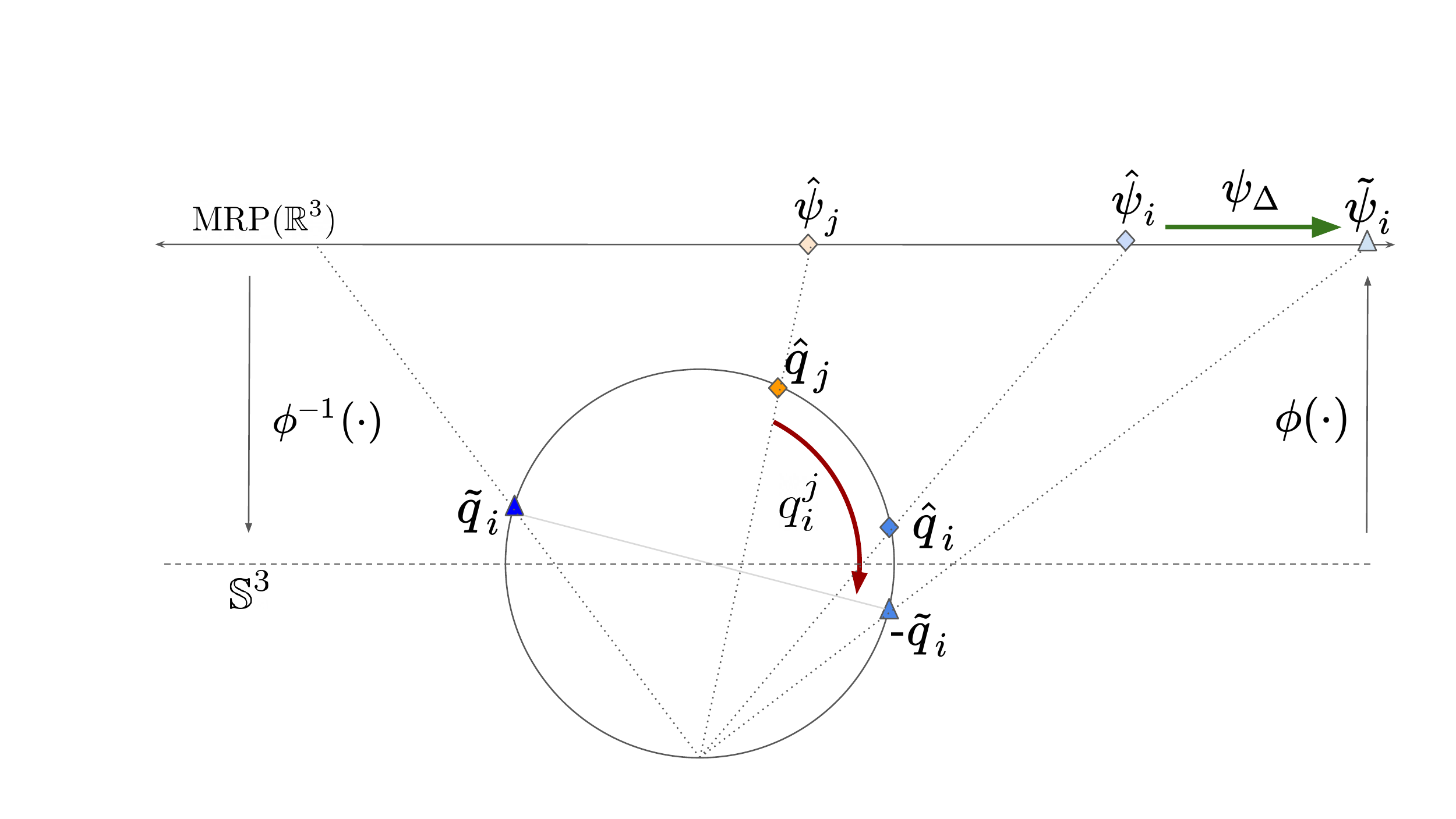}
 	\vspace{-1em}
 	\caption{Projection of relative supervision, $q_i^j$, shown in \textcolor{BrickRed}{red}, between rotations $\hat{q}_j:=\phi^{-1}(\hat{\psi}_j)$ and $-\tilde{q}_i$, into the MRP space update, $\psi_\Delta$, shown in \textcolor{PineGreen}{green}. While $\tilde{q}_i$ could have been selected as the the goal rotation, it would have induced a much larger movement in the projected space.}
 	\label{fig:mrp_delta_projection}
\end{wrapfigure}

As mentioned 
previously, optimizing on a closed space, such as $SO(3)$ or $\mathbb{S}^3$ can be problematic, since the relative distance between two points can eventually be minimized by moving them in the exact opposite direction of the minimum path between them. 
To alleviate this issue, we would like to instead perform self-supervised learning in an open space, where this symmetry is broken. 
This can be done using Modified Rodrigues Parameters (MRP)~\cite{wiener1962theoretical,terzakis2018modified}. MRP is the stereographic projection of the closed manifold of the quaternion sphere $\mathbb{S}^{3}$ to $\mathbb{R}^3$, and has been widely used in attitude estimation and control \cite{crassidis1996attitude}. 
In combining this projection with the mapping between $SO(3)$ and $\mathbb{S}^3$, this projection can be used to optimize rotations. 
We define a unit quaternion $q = \begin{bmatrix} \rho & \nu \end{bmatrix}\in \mathbb{S}^3\triangleq \{x\in \mathbb{R}^{4}: \|x\| = 1\}$, where $\rho \in \mathbb{R}$ defines the scalar component and $\nu \in \mathbb{R}^3$ defines the imaginary vector component of the unit quaternion.  The projection operator $\phi(q) = \psi \in \mathbb{R}^3$ and its inverse $\phi^{-1}(\psi) = q \in \mathbb{S}^3$ are given by~\cite{wiener1962theoretical,terzakis2018modified}
where $\psi = \phi\left(\begin{bmatrix}\rho & \nu \end{bmatrix}\right) = \frac{\nu}{1+\rho}$ and $\begin{bmatrix}\rho & \nu \end{bmatrix} = \phi^{-1}(\psi) = \begin{bmatrix} \frac{1-\|\psi\|^2}{1+\|\psi\|^2} & \frac{2 \psi}{1+\|\psi\|^2} \end{bmatrix}$.
Given this projective orientation space, we need to map our relative rotation $R_i^j$ into the projective space in order to use these relative rotations for the self-supervised learning task. 


This projection is required, as the relative supervision is in $SO(3)$, and the direction and magnitude of this relative measurement are distorted differently in different regions of the projective MRP space. 
Given a pair of estimated projected rotations $\hat{\psi}_i := \phi(\hat{q}_i)$ and $\hat{\psi}_j := \phi(\hat{q}_j)$, 
we project $\hat{\psi}_j$ back to a unit quaternion $\phi^{-1}(\hat{\psi}_j) = \hat{q}_j \in \mathbb{S}^3$ and rotate it according to $R_i^j$, $\tilde{q}_i = q_i^j \otimes \hat{q}_j$, where $\otimes$ is quaternion multiplication and $q_i^j$ is the quaternion form of $R_i^j$. The resulting unit quaternion $\tilde{q}_i$ is then projected back into the Modified Rodrigues Parameter space, $\tilde{\psi}_i$. 
A simplified visual analogy of this process is shown in Figure~\ref{fig:mrp_delta_projection}. 

While this relative rotation could be applied and projected at either the sampled point $\hat{\psi}_i$, or the neighboring location $\hat{\psi}_j$, we select the neighboring location $\hat{\psi}_j$, as it does not require us to compute gradients through the forward or inverse projections $\phi(\cdot)$ and $\phi^{-1}(\cdot)$, respectively.
This projected rotation $\tilde{\psi}_i$ represents the value $\hat{\psi}_i$ should hold, relative to the current predicted rotation $\hat{\psi}_j$. It should be noted that $\phi(q) \neq \phi(-q)$, while $q$ and $-q$ represent the same rotation. In terms of the projective space, this means that the sign of $\tilde{q}_i$ matters. To remove this ambiguity, we select the nearest projection to $\hat{\psi}_i$ in the projective MRP space.
It should be noted that this is different from selecting the closer antipode on $\mathbb{S}^3$, as the large deformations found near the south pole\footnote{The south pole in this case is described by the quaternion $-1 + 0i + 0j + 0k$} can cause the nearer antipode in $\mathbf{S}^3$ to be further in MRP space. In contrast, if we were to select a consistent sign for the scalar component $\tilde{q}_i$, for example ensuring the scalar component is always positive, a small change in $\hat{\psi}_j$ can cause large changes in $\tilde{\psi}_i$ when $\phi^{-1}(\hat{\psi}_j)$ is near the equator of $\mathbb{S}^3$. 
While this change is required to stabilize our optimization, it does add some ambiguity to the direction of optimization. However, the directions to each of the projected locations, $\phi(\tilde{q}_i)$ and $\phi(-\tilde{q}_i)$,  can only be anti-parallel (pulling in exactly opposite directions) if $\tilde{\psi}_i - \hat{\psi}_i$ intersects the origin in $\mathbb{R}^3$. 

The loss with respect to a given estimate, $\hat{\psi}_i$, can then be written as the $l_2$ distance between its current value and the projected relative location, $\tilde{\psi}_i$, relative to a given neighbor, $\hat{\psi}_j$:
\begin{subequations}
\begin{minipage}{.33\textwidth}
    \begin{equation} 
        \mathcal{L}_{\Psi+} = \left\|\hat{\psi}_i - \phi(\tilde{q}_{i})\right\|^2
    \end{equation}
\end{minipage}%
\begin{minipage}{.33\textwidth}
    \begin{equation} 
        \mathcal{L}_{\Psi-} = \left\|\hat{\psi}_i - \phi(-\tilde{q}_{i})\right\|^2
    \end{equation}
\end{minipage}
\begin{minipage}{.33\textwidth}
    \begin{equation} 
        \mathcal{L}_{\Psi} = \min(\mathcal{L}_{\Psi-}, \mathcal{L}_{\Psi+})
    \end{equation}
\end{minipage}
\label{eq:mrp_loss}
\end{subequations}

where we recall that, $\tilde{q}_i = q_i^j \otimes \hat{q}_j$, and $\hat{q}_j = \phi^{-1}(\hat{\psi}_j)$.

Note that, while $\hat{\psi}_j$ is a predicted value, we do not pass gradients through it, allowing it to anchor the update to a consistent orientation. The gradient update\footnote{We omit a constant factor for brevity, and integrate it into the learning rate, $\gamma$.} is then given by:
\begin{equation}
    \nabla_{\hat{\psi}_i} \mathcal{L}_{\Psi} = \psi_\Delta = \begin{cases}
 \hat{\psi}_i - \phi\left(\tilde{q}_{i}\right), & \text{if }\mathcal{L}_{\Psi+} < \mathcal{L}_{\Psi-} \\
 \hat{\psi}_i - \phi\left(-\tilde{q}_{i}\right), & \text{otherwise}
\end{cases}
\label{eq:mrp_grad}
\end{equation}
Additionally, a maximum gradient step, $\eta$, in the projective space is imposed,
$\psi_\Delta \gets \eta \frac{\psi_\Delta}{\|\psi_\Delta\|}$, if the gradient exceeds a defined amount.
This prevents extremely large steps from being taken, as the projective transform can distort the space. 

\section{Experiments}
\label{sec:experiments}
Next, we perform experiments to show that our method converges faster and more consistently than the alternative approaches. Our empirical results are grouped into two settings: (1) direct optimization of randomly generated rotations, Section~\ref{sec:direct_optimization}, and (2) optimization of the parameters of a convolutional neural network using synthetically rendered images, Section~\ref{sec:network_optimization}. In both cases, relative orientations between elements in a neighborhood are provided. We show Iterative Modified Rodrigues Projective Averaging is able to converge faster and more often than alternative approaches. We further show in Section~\ref{sec:network_optimization} that our method can easily be used to supervise convolutional neural networks, when only relative orientation information is available. 

\begin{figure}[t]
 	\centering
 	\begin{subfigure}{0.45\textwidth}
        \centering
     	\includegraphics[width=\textwidth, trim={0 0 0 0},clip]{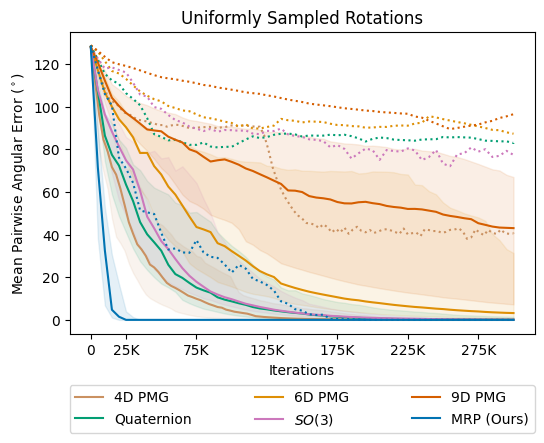}
 	\end{subfigure}
    \begin{subfigure}{0.45\textwidth}
        \centering
     	\includegraphics[width=\textwidth]{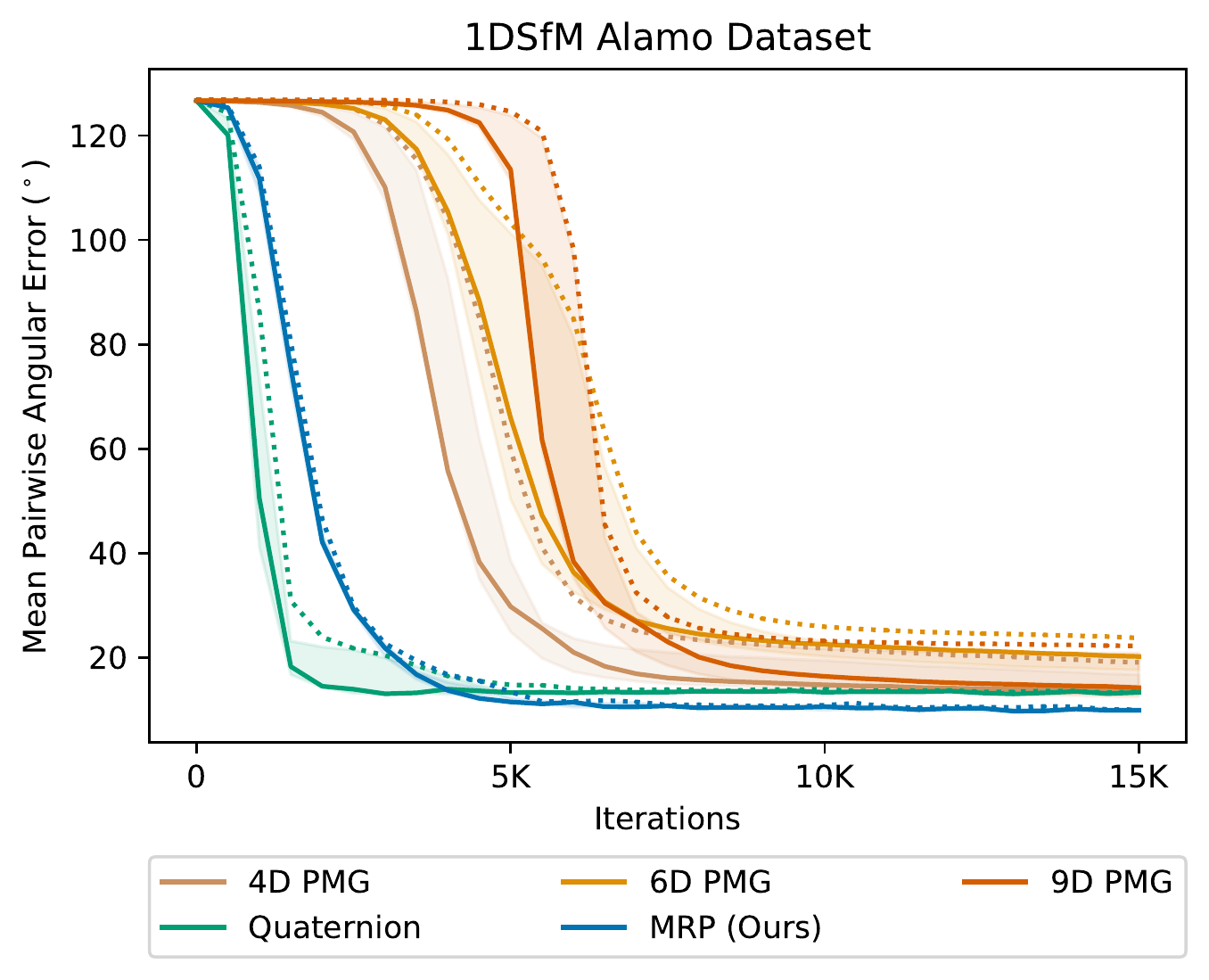}
    \end{subfigure}
 	\caption{Relative rotation consensus with direct optimization of rotation parameters over 50 unique environments with 100 random generated orientations each (\textbf{left}) and Alamo 1DSfM~\cite{wilson_eccv2014_1dsfm} (\textbf{right}).
 	 Median average-pair-wise angular error (${}^\circ$) between each estimated rotations is shown, with shaded region representing the first and third quartile for each method. The max average-pair-wise angular error for each algorithm at each iteration is shown as a dashed line.} 
 	 
 	\label{fig:direct_convergence}
\end{figure}
\begin{table}[t]
\footnotesize
    \centering
    \begin{tabular}{c|c c c|c c c}
    \hline
     & \multicolumn{3}{c}{\textit{Avg Pairwise Angular Error $< 5^{\circ}$}} & \multicolumn{3}{|c}{\textit{Normalized AUC}} \\
     Algorithm & Mean Steps & Max Steps & Min Steps & Mean & Max & Min \\
    \hline
    \hline
        $SO(3)$ & 157.7K & Not Converged & 85.0K & 24.47 & 82.92 & 7.55 \\
        4D PMG~\cite{chen2021projective} & 126.1K & Not Converged & 27.0K & 15.67 & 52.40 & 3.06 \\
        6D PMG~\cite{zhou2019continuity} & 235.9K & Not Converged & 80.0K & 43.53 & 89.15 & 11.34 \\
        9D PMG~\cite{levinson2020analysis} & 284.5K & Not Converged & 150.0K & 62.94 & 101.77 & 17.77 \\
        Quaternion & 160.3K & Not Converged & 40.0K & 23.55 & 84.85 & 3.47 \\
        \textbf{MRP (Ours)} & \textbf{37.5K} & \textbf{160.0K} & \textbf{15.0K} & \textbf{5.08} & \textbf{15.56} & \textbf{2.18} \\
    \hline
    \end{tabular}
    \caption{Number of iteration steps until convergence and Normalized Area Under Curve (nAUC) over 50 unique environments of 100 randomly generated orientations. 300K optimization steps are taken for each experiment.}
    \label{tab:dirrect_converge}
     	\vspace{-1em}

\end{table}
\begin{table}[b]
\footnotesize
    \centering
    \begin{tabular}{c|c c c c c|c c }
    \hline
        & \multicolumn{5}{c}{\textit{\% Avg Pairwise Angular Error $ < 5^{\circ}$}} & \multicolumn{2}{|c}{\textit{Final Error(${}^{\circ}$)}} \\
    Algorithm & 30K  & 70K  & 100K  & 150K  & 300K & Mean & Median \\
    \hline
    \hline
    $SO(3)$ & 0\% & 0\% & 6\% & 57\% & 94\% & 2.056 & 0.10 \\
    4D PMG~\cite{chen2021projective} & 2\% & 32\% & 46\% & 72\% & 90\% & 1.969 & 0.14 \\
    6D PMG~\cite{zhou2019continuity} & 0\% & 0\% & 4\% & 20\% & 52\% & 20.096 & 3.20 \\
    9D PMG~\cite{levinson2020analysis} & 0\% & 0\% & 0\% & 2\% & 20\% & 40.125 & 43.02 \\
    Quaternion & 0\% & 12\% & 30\% & 56\% & 82\% & 9.72 & 0.04 \\
    \textbf{MRP (Ours)} & \textbf{66\%} &\textbf{88\%} & \textbf{96\%} & \textbf{98\%} & \textbf{100\%} & \textbf{0.004} & \textbf{0.004} \\

    \hline
    \end{tabular}
    \caption{Percentage of experiments converged and final angular errors over 50 unique environments of 100 randomly generated orientations. 300K optimization steps are taken for each experiment.}
    \label{tab:direct_errors}
 	\vspace{-1em}
\end{table}


\subsection{Direct Parameter Optimization}
\label{sec:direct_optimization}
 \looseness=-1
We evaluate the convergence behaviour of our Iterative Modified Rodrigues Projective Averaging method, \textbf{MRP (Ours)} , described in Section~\ref{sec:mrp_consensus}, as well as the $SO(3)$ averaging method, described in Section~\ref{sec:so3_consensus}. 
For the $SO(3)$ averaging method, we implement both the pure Riemannian optimization,
 $\mathbf{SO(3)}$, as well as a method using a Projective Manifold Gradient Layer~\cite{chen2021projective} to map the Riemannian gradient of the $SO(3)$ averaging loss, Equation~\ref{eq:so3_loss}, to a Euclidean optimization in $\mathbb{R}^D$, where we test $D=4,6$, and $9$, \textbf{4D PMG~\cite{chen2021projective}}, \textbf{6D PMG~\cite{zhou2019continuity}}, \textbf{9D PMG~\cite{levinson2020analysis}}, respectively.
Additionally, we evaluate direct quaternion optimization, described in Sections~\ref{sec:quat_consensus}, \textbf{Quaternion}.

\looseness=-1
\textbf{Uniformly Sampled Rotations.}
\label{sec:uniform_rotations}
We test the performance of each algorithm when directly optimizing the rotation parameters of a set of size $N=100$ with known relative rotations $R_i^j$, and local neighborhood structure. Ground truth and initial estimated rotations 
are both randomly sampled from a uniform distribution in $SO(3)$. Each rotation, $R_i$, has a neighborhood, $\mathcal{N}_i$, consisting of the closest $|\mathcal{N}_i|=3$ rotations with respect to geodesic distance. The connectivity of this neighborhood graph is checked to ensure the graph contains only a single connected component. We test all algorithms over 50 sets of unique environments, each with $N=100$ randomly generated orientations as described above. The estimated rotations are updated by each algorithm in batches of size 8, for 300K iterations. 

As the goal of our algorithm is to improve the convergence properties of iterative averaging methods, we analyze each algorithm at various stages of optimization. We are particularly interested in the average number of update steps until the algorithm has converged, which we define as when the average angular error between all pairs of rotations is below $5^\circ$. 
As we can see in Figure~\ref{fig:direct_convergence}, the Iterative Modified Rodrigues Projective Averaging method, \textbf{MRP (Ours)}, converges before the standard $SO(3)$ averaging method.
On average, our method converged to within $5^\circ$ in 37K steps. The next best method, \textbf{4D PMG~\cite{chen2021projective}}, which takes over three times as many iterations to converge to the same level of accuracy. Further, Table~\ref{tab:dirrect_converge} shows that our method is the only one to converge across all environments within 300K iterations. 
For each method, we also compute the mean area under the pairwise error curve, with the number of steps normalized to between zero and one (nAUC), also shown in Table~\ref{tab:dirrect_converge}. We find that in the best, average, and worst case scenarios, our method has the best convergence behavior.
To quantify convergence behavior, we also compute the percentage of trials that achieve average pairwise angular error below $5^{\circ}$ at different stages of training, as shown on the left in Table~\ref{tab:direct_errors}.
We find that at each stage of training, the Iterative Modified Rodrigues Projective Averaging, \textbf{MRP (Ours)}, training has a lower average pairwise error, shown in Table~\ref{tab:direct_errors}. Our method also converged far more often at each stage of training, also shown in Table~\ref{tab:direct_errors}.



\textbf{Structure from Motion Dataset.}
To test our algorithms under natural noise conditions, we also evaluate our algorithm on the 1DSfM~\cite{wilson_eccv2014_1dsfm} structure from motions datasets. 
Each environment is tested with 5 random initializations and the estimated rotations are updated by each algorithm in batches of size 64, for 20K iterations. The results of a subset of the environments are shown in Table~\ref{tab:1dsfm_results} and the remainder can be found in Appendix~\ref{appendix:1dsfm_dataset}. 
The noise characteristics of relative rotations in this dataset are similar to those found when capturing relative poses, but, unlike the Uniformly Sampled Rotations environments,
the distribution of rotations does not fully cover the orientation space. 
As a result, all methods converge relatively quickly.
Our algorithm outperforms the baselines in terms of accuracy.  While the \textbf{Quaternion} optimization converges slightly faster, it consistently finds a lower accuracy configuration, resulting in a low nAUC, but higher relative and absolute accuracy. 
More details can be found in the Appendix~\ref{appendix:1dsfm_dataset}. 

\begin{table}[h]
\footnotesize
    \centering
    \begin{tabular}{c|c c c c c c}
    \hline
     & \multicolumn{2}{c}{\textit{Mean Relative}} & \multicolumn{2}{c}{\textit{Mean Absolute}} & \multicolumn{2}{c}{} \\
     & \multicolumn{2}{c}{\textit{ Error} (${}^{\circ}$)} & \multicolumn{2}{c}{\textit{ Error} (${}^{\circ}$)} & \multicolumn{2}{c}{\textit{Mean nAUC}} \\
    Algorithm & E. Island & Alamo & E. Island & Alamo & E. Island & Alamo \\
    \hline
    4D PGM~\cite{chen2021projective} &  11.94 &  15.00 &  7.34 &  9.94 &  25.60 &  47.20 \\ 
    6D PGM~\cite{zhou2019continuity} &  11.26 &  18.84 &  6.90 &  13.09 &  27.77 &  58.04 \\ 
    9D PGM~\cite{levinson2020analysis} &  10.22 &  16.32 &  6.32 &  11.43 &  29.31 &  60.14 \\ 
    Quaternion &  11.58 &  13.40 &  7.23 &  8.93 & \textbf{ 16.01 } &  \textbf{22.57} \\ 
    \textbf{MRP (Ours)} &  \textbf{8.84} &  \textbf{9.89} &  \textbf{5.49} &  \textbf{6.56} &  16.21 &  25.61 \\
    \hline
    \hline
    IRLS-GM\cite{chatterjee2013efficient} & - & - & 3.04 &  3.64  & - &  - \\
    IRLS-$\ell_\frac{1}{2}$\cite{chatterjee2017robust} & - & - & 2.71 &  3.67  & - &  - \\
    MLP\cite{shi2020message} & - & - & 2.61 &  3.44  & - &  - \\
    \hline
    \end{tabular}
    \caption{\textbf{Rotation Averaging Results on 1DSfM ~\cite{wilson_eccv2014_1dsfm} dataset.} Results before the double lines are comparisons of local method 
    after 20K iterations. Results under the double lines are obtained from global methods which are incompatible with SGD training. 
    }
    \label{tab:1dsfm_results}
\end{table}
\begin{wrapfigure}[13]{r}{0.55\textwidth}
\capbtabbox{

\footnotesize
    \centering
    \begin{tabular}{c|c c c}
    \hline
     & Mean & Median & \\
    Algorithm & Error (${}^{\circ}$) & Error (${}^{\circ}$) & $5^{\circ}$ Acc (\%)\\
    \hline
     4D PMG~\cite{chen2021projective} &  123.84 & 123.96 & 0 \\
     Quaternion & 28.83 & 21.74 &  50 \\
     \textbf{MRP (Ours)} & \textbf{3.71} & \textbf{3.73} & \textbf{100}\\
     \hline\hline
     Oracle  & 1.58 & 1.56 & 100 \\
    \hline
    \end{tabular}
}{
    \captionof{table}{\textbf{Final results for image based rotation estimation.} Final mean and median angular error  (${}^{\circ}$) and percentage of seeds below $5^{\circ}$ after 10K steps over 8 unique environments of 100 random rotations. 
    } 
    \label{tab:nn_stats}
}

\end{wrapfigure}

\subsection{Neural Network Optimization}
\label{sec:network_optimization}

To show that the Iterative Modified Rodrigues Projective Averaging method, \textbf{MRP (Ours)}, 
can be used to learn orientation using neural networks by optimizing the parameters of a simple CNN, specifically a ResNet18~\cite{he2016deep}, we follow the procedure as in Section~\ref{sec:direct_optimization} with some minor changes. Instead of operating directly on a set of rotation parameters, we learn a function $\hat{\psi}_i = f(I_i)$ from rendered images of the YCB drill~\cite{xiang2018posecnn} model
rendered at each of 100 random orientations $R_i$. We continue to only supervise each method described in Section~\ref{sec:direct_optimization} using the relative rotations between each image. 

We compare the best performing methods, and, as a lower bound, we also train an oracle network, \textbf{Oracle}, with the ground truth rotations, $R_i$ and cosine quaternion loss.
We use a batch size of 32 and the Adam ~\cite{kingma2015adam} optimizer with a learning rate of $10^{-4}$ for all experiments. All methods are trained for a maximum of 10K steps, over 8 environments, each with 100 images of randomly generated rotations. We report final mean and median pairwise angular error, and the percentage of runs converged below $5^{\circ}$ pairwise angular error as $5^{\circ}$ Acc. We find that \textbf{MRP (Ours)} is able to converge to a rotational frame
consistent with the relative rotations used for supervision relatively quickly, with a significantly lower average-pairwise-error than all other relative methods, shown in 
Table~\ref{tab:nn_stats}.  

We perform experiments for generalization to unseen poses and find that a curriculum is required (see Appendix~\ref{appendix:curriculum} for details).  For the generalization experiments,
we found that \textbf{MRP (Ours)} achieves a mean pairwise angular error or $5.19^{\circ}$, \textbf{Quaternion} achieves $12.41^{\circ}$, and \textbf{
4D PMG~\cite{chen2021projective}} never converged, with final error of $125.09^{\circ}$. 


\subsection{3D Object Rotation Estimation via Relative Supervision from Pascal3D+ Images}
\label{sec:pascal_exp}
\textbf{Experimental Setup.} Pascal3D+~\cite{xiang2014beyond} is a standard benchmark for categorical 6D object pose estimation from real images. 
Following~\cite{chen2021projective, levinson2020analysis}, we discard occluded or truncated objects and augment with rendered images from~\cite{su2015render}. We report 3D object pose estimation results on two object categories:
\textit{sofa} and \textit{bicycle}. We compare our method \textbf{MRP} with five baselines: \textbf{Quaternion}, \textbf{4D PMG}~\cite{chen2021projective}, \textbf{6D PMG} ~\cite{zhou2019continuity}, \textbf{9D PMG} ~\cite{levinson2020analysis} and \textbf{10D PMG}~\cite{peretroukhin2020smooth}, all of which use ResNet18~\cite{he2016deep} backbone to predict the object representation. Each model is supervised using the geodesic error between the relative orientation of the predicted absolute orientations, and the relative orientation between the ground truth absolute orientations, for each image pair. 
We use the same batch size of 20 as in~\cite{levinson2020analysis,chen2021projective}, and use Adam~\cite{kingma2015adam} with learning rate of $10^{-4}$.
 
\textbf{Result Analysis.} Results for \textit{sofa} are showed in 
Table~\ref{fig:pascal_sofa_merged}(a); \textit{bicycle} are showed in Table~\ref{fig:pascal_sofa_merged}(b).
For \textit{sofa} category, we find that after 80K training iterations, \textbf{MRP (Ours)} achieves a mean angular pairwise error of $13.63^{\circ}$ on the test set, outperforms all other baselines. \textbf{10D PMG} achieves the worst error out of all methods, with final angular pairwise error of $19.28^{\circ}$. 
For \textit{bicycle} category, we find that after 80K training iterations, \textbf{MRP (Ours)} achieves a mean angular pairwise error of $29.46^{\circ}$ on the test set, outperforms all other baselines.  In both the \textit{sofa} and the \textit{bicycle} category, we find that $\textbf{MRP (Ours)}$ has the fastest convergence speed, in addition to achieving the lowest test angular error. More details can be found in Appendix~E.
 
 
 


\begin{figure}[!t]

\subfloat[]{%
\scriptsize
\refstepcounter{table}
\label{tab:actionK-HTFETI-crossVal}
\begin{tabular}{c|c c c}
 
    \hline
    & \multicolumn{3}{c}{\textit{Final Test Angular Pairwise Error}}\\
     Algorithm  &  Mean (${}^{\circ}$) & Max (${}^{\circ}$) & Min (${}^{\circ}$)\\
    \hline
     4D PMG~\cite{chen2021projective} & 17.39 $\pm$ 1.14 & 19.42 & 16.07\\
     6D PMG~\cite{chen2021projective, zhou2019continuity} &  15.20 $\pm$ 0.77 & 16.43 & 14.44 \\
     9D PMG~\cite{chen2021projective, levinson2020analysis} &  14.61 $\pm$ 0.50 & 15.66 & 14.18\\
     10D PMG~\cite{chen2021projective, peretroukhin2020smooth} & 19.28 $\pm$ 7.58 & 37.76 & 15.03\\ 
     Quaternion &  16.52 $\pm$ 4.12 & 26.57 & 14.38\\ 
     \textbf{MRP (Ours)} &  \textbf{13.63 $\pm$ 0.78} & \textbf{15.08} & \textbf{12.62}\\
    \hline
    \end{tabular}
}
\quad 
\subfloat[]{%
\scriptsize
\refstepcounter{table}
\label{tab:actionK-HTFETI-crossVal}
    \begin{tabular}{c|ccc}
    \hline    
    & \multicolumn{3}{c}{\textit{Final Test Angular Pairwise Error}}\\
     Algorithm  &  Mean (${}^{\circ}$) & Max (${}^{\circ}$) & Min
     (${}^{\circ}$) \\
    \hline
     4D PMG~\cite{chen2021projective} & 34.57 $\pm$ 2.21 & 38.13 & 31.90 \\
     6D PMG~\cite{chen2021projective, zhou2019continuity} &  31.58 $\pm$ 2.24 & 35.66 & \textbf{28.42}\\
     9D PMG~\cite{chen2021projective, levinson2020analysis} &  31.80 $\pm$ 1.52 & 34.87 & 29.96  \\
     10D PMG~\cite{chen2021projective, peretroukhin2020smooth} &  32.23 $\pm$ 2.10 & 36.98 & 29.87 \\ 
     Quaternion &  31.92 $\pm$ 1.00 & 33.61 & 30.61\\ 
     \textbf{MRP (Ours)} &  \textbf{29.46 $\pm$ 0.66} & \textbf{30.74} & 28.62\\

    \hline
    \end{tabular}
}
 
\captionof{table}{\textbf{3D Object Pose Estimation via Relative Supervision on Pascal3D+ test images for \textit{sofa}, \textit{bicycle}}. 
Final mean test angular pairwise error on Pascal3D+ \textit{sofa} (a), \textit{bicycle} (b) images after 80K training iterations, over 8 random seeds.}
\label{fig:pascal_sofa_merged}
\end{figure} 

  




\section{Conclusion and Limitations}
\looseness=-1
In this paper, we show that through the use of Modified Rodrigues Parameters, we are able to open the closed manifold of $SO(3)$, improving the convergence behavior of the rotation averaging problem. 
Additionally, we show that our method, Iterative Modified Rodrigues Projective Averaging, is able to outperform the naive application of relative-orientation supervision in both 
direct parameter optimization and image-based rotations estimation from neural networks.   
While this parameterization 
is valuable for learning rotations through relative supervision, it is not without limitations. One of the primary ones is the need for a curriculum to generalizing to unseen relative rotations. Without this, our experiment show that all representations fall into the local optima of predicting a constant orientation. Additionally, in the generalization experiments, we were only able to achieve a final error of 5 degrees, which may not be accurate enough for many fine motor tasks. 
We hope our method allows more systems to convert the relative supervision of relative methods, like ICP, to consistent and accurate absolute poses.

\section{Acknowledgements}
We thank Prof Zachary Manchester in particulate for pointing us to Modified Rodriguez Parameters. Additionally, we thank 
Prof Frank Dellaert, Mark Gillespie, Prof Richard Hartley, Valentin Peretroukhin, Prof David Rosen, Prof Denis Zorin, and Prof Keenan Crane for taking time to meet with us and providing enormously helpful feedback and advice. This work is supported by the National Science Foundation under Grant No. IIS-1849154 and by LG Electronics.

\bibliography{root}

\begin{bibunit}[corlabbrvnat]
\appendix
\newpage

\renewcommand{\thesection}{\arabic{section}}
\appendix


\etocsetlevel{appendixplaceholder}{-1}
\etoctoccontentsline*{appendixplaceholder}{APPENDIX}{-1}

\title{Deep Projective Rotation Estimation through Relative Supervision - Appendix}

\maketitle

\localtableofcontents
\appendix

\section{Intuitive Example}
\label{appendix:theory}
We present an intuitive example of when optimizing a set of orientations to solve the rotation averaging problem described in Equation~(1), in the main text, 
can fail. In this example, we show the benefits of the Iterative Modified Rodrigues Projective Averaging approach over the baseline approach. We show that, while both $SO(3)$ averaging and Iterative Modified Rodrigues Projective Averaging share a class of non-optimal critical points, in the projective case, these critical points are a subset of the problematic configurations for $SO(3)$ averaging. 

\subsection{Examples of Critical Points}
\label{appendix:theory-so3}
In this section, we analyze a class of critical points shared by both standard $SO(3)$ averaging and Iterative Modified Rodrigues Projective Averaging. For simplicity, we will examine the $N=3$ rotation case, where $\mathcal{R} = \{R_1, R_2, R_3\}$ with relative rotations of $R_i^j := R_i R_j^\top$. 
As this is an iterative algorithm, we need to initialize our predicted rotations to some values $\hat{\mathcal{R}} = \{\hat{R}_1, \hat{R}_2, \hat{R}_3\}$. In this case, we initialize each predictions to  $\hat{R}_i := R_i R_0 \exp\left(\left(\theta_0 + i\frac{2 \pi}{N}\right) \omega_0\right)$ 
where $R_0$ is an arbitrary but constant rotational offset, $\omega_0$ and $\theta_0$ define an arbitrary, but constant axis and constant rotation, about which each initial estimate $\hat{R}_i$ is rotated an additional angle of $\theta_i$.
We find that if we use the previously described methods to update this initial configuration, under certain values of $\mathcal{R}$ , $R_0$, $\theta_0$, and $\omega_0$, the expected update at each value $\hat{R}_i$ is $\mathbf{0}$, forming a critical point for each algorithm.

\subsubsection{Critical Point for $SO(3)$ Averaging}
\label{appendix:so3_optima}
Given the initial predictions of $\hat{\mathcal{R}}$ defined above, for all values of $\mathcal{R}$ , $R_0$, $\theta_0$, and $\omega_0$, we find that the expectation of the gradient of $SO(3)$ averaging loss, $\mathbb{E}_{i,j}\left[\nabla_{\hat{r}_i} \mathcal{L}_{SO(3)}\right]$, is $\mathbf{0}$. 
The gradient of any sampled pair $i,j$ is given by
\begin{align*}
    \nabla_{i} \mathcal{L}_{SO(3)}^{i,j}:&= \nabla_{\hat{r}_i} \mathcal{L}_{SO(3)} \left(\hat{R}_i, \hat{R}_j, R_i^j\right) \\
    &= \log \left(\hat{R}_i^\top R_i^j \hat{R}_j\right) \\
    &= \log\left(\left(R_i R_0 \exp\left(\theta_i \omega_0\right)\right)^\top R_i^j R_j R_0 \exp\left(\theta_j \omega_0\right)\right) \\
    &= \log\left(\exp\left((\theta_j - \theta_i) \omega_0\right)\right) \\
    &= \text{wrap}_{[-\pi, \pi)}\left[(\theta_j - \theta_i) \right] \omega_0 \\
    &= \text{wrap}_{[-\pi, \pi)}\left[\frac{2 \pi}{N}(j - i) \right] \omega_0 \\
    &= \frac{2 \pi}{N}(j - i) \omega_0.
\end{align*}
This lead to an expected gradient of each estimate rotation $\hat{R}_i$ of 

\begin{equation*}
    \mathbb{E}_{j}\left[\nabla_{\hat{r}_i} \mathcal{L}_{SO(3)} \left(\hat{R}_i, \hat{R}_j, R_i^j \right) \Big|i=1 \right]
    = \frac{1}{2}  \text{wrap}_{[-\pi, \pi)}\left[ \sum_{j \neq i}  \frac{2 \pi}{N}(j - i) \right] \omega_0 
    = \mathbf{0}.
\end{equation*}

For all estimates $\hat{R}_i$, this sums to an integer multiple of $2 \pi \omega_0$, which, due to the definition of the $SO(3)$ exponential map, wraps to $\mathbf{0}$. 



\subsubsection{Critical Point for Iterative Modified Rodrigues Projective Averaging}
\label{appendix:mrp_optima}
When optimizing using our Iterative Modified Rodrigues Projective Averaging method, we find that this configuration is only a critical point when the relative orientations between each pair of rotations are equal and opposite, i.e., $R_i^j = R_i^{k\top} \to R_i^j = \exp\left(\pm \frac{2 \pi}{N} \omega_0\right)$ and the predicted orientations are initialized at identity: $R_0 = \mathbf{I}$. This only happens when the true orientations $\mathcal{R}$ are evenly spaced about an axis of rotations: $R_i:= \exp\left(\left(\theta_0 - i\frac{2 \pi}{N}\right) \omega_0\right)$, leaving only axis of rotation $\omega_0$ and the constant angular offset $\theta_0$ about that axis as free parameters. 

As we are trying to update these rotations using a method compatible with stochastic gradient descent, we are concerned with the expectation of our update with respect to a sampled pair. In this case, the expected loss and update, defined in Equations~4c 
and~5 in the main paper, respectively, for any projected rotation $\hat{\psi}_i$ and its neighbor $\hat{\psi}_j$ is $\mathcal{L}_{\Psi+}^{i,j}:= \left\|\hat{\psi}_i - \phi(q_i^j \otimes \phi^{-1}(\hat{\psi}_j))\right\|^2$ where $q_i^j$ is the quaternion associated with $R_i^j$.
As all $\hat{\psi}_i$ are initialized to the identity, i.e., $\phi(q_I) = \mathbf{0}$ where $q_I$ is the identity quaternion, we get

\begin{subequations}
\begin{minipage}{.5\textwidth}
    \begin{equation*} 
    \mathcal{L}_{\Psi+}^{i,j} := \left\|-\phi^{-1}(q_i^j)\right\|^2 \\
    \end{equation*}
\end{minipage}%
\begin{minipage}{.5\textwidth}
    \begin{equation*} 
    \nabla_{i} \mathcal{L}_{\Psi+}^{i,j}:= -\phi^{-1}(q_i^j) \\
    \end{equation*}
\end{minipage}
\end{subequations}

\begin{subequations}
\begin{minipage}{.5\textwidth}
    \begin{equation*} 
    \mathcal{L}_{\Psi-}^{i,j}:= \left\|-\phi^{-1}(-q_i^j)\right\|^2 \\
    \end{equation*}
\end{minipage}%
\begin{minipage}{.5\textwidth}
    \begin{equation*} 
    \nabla_{i} \mathcal{L}_{\Psi-}^{i,j}:= -\phi^{-1}(-q_i^j) 
    \end{equation*}
\end{minipage}
\end{subequations}

The relative rotations in this configuration are 
\begin{equation*}
R_i^j := \exp\left(\pm \frac{2 \pi}{3} \omega_0\right)
\end{equation*}
with relative quaternions $q_i^j := \begin{bmatrix} \cos(\frac{\pi}{3}) & \pm \sin(\frac{\pi}{3}) \omega_0 \end{bmatrix},
$ which leads to 

\begin{subequations}
\begin{minipage}{.5\textwidth}
    \begin{equation*} 
        \phi(q_i^j) = \frac{\pm \sin(\frac{\pi}{3}) \omega_0}{1 + \cos(\frac{\pi}{3})} = \frac{\pm \omega_0}{\sqrt{3}}
    \end{equation*}
\end{minipage}%
\begin{minipage}{.5\textwidth}
    \begin{equation*} 
        \phi(-q_i^j) = \frac{\mp \sin(\frac{\pi}{3}) \omega_0}{1 - \cos(\frac{\pi}{3})} = \pm \sqrt{3} \omega_0.
    \end{equation*}
\end{minipage}
\end{subequations}

This results in the potential losses for the positive and negative antipodes of 

\begin{subequations}
\begin{minipage}{.5\textwidth}
    \begin{equation*} 
        \mathcal{L}_{\Psi+}^{i,j} = \|\phi(q_i^j)\| = \frac{1}{3}
    \end{equation*}
\end{minipage}%
\begin{minipage}{.5\textwidth}
    \begin{equation*} 
        \mathcal{L}_{\Psi-}^{i,j} = \|\phi(-q_i^j)\| = 3
    \end{equation*}
\end{minipage}
\end{subequations}

for all pairs of $i,j$. Selecting the minimum loss antipodes, we get gradients of

\begin{subequations}
\begin{minipage}{.5\textwidth}
    \begin{equation*} 
    \nabla_{i}\mathcal{L}_{\Psi}^{i,j} = \frac{\mp 1}{\sqrt{3}} \omega_0
    \end{equation*}
\end{minipage}%
\begin{minipage}{.5\textwidth}
    \begin{equation*} 
 \nabla_{i}\mathcal{L}_{\Psi}^{i,j} = \frac{\pm 1}{\sqrt{3}} \omega_0,
    \end{equation*}
\end{minipage}
\end{subequations}
for $j = i+1$ and $j = i-1$, respectively. The final expectation of the gradients with respect neighborhood sampling is
\begin{align*}
    \mathbb{E}_{j}\big[\nabla_{\hat{\psi}_i} \mathcal{L}_{SO(3)}(\hat{\psi}_i, \hat{\psi}_j, R_i^j) |i=1\big] 
    = \frac{1}{2} \sum_{j \neq i}  \nabla_{i} \mathcal{L}_{\Psi}^{i,j} 
    = \frac{1}{2} \left(\frac{1}{\sqrt{3}} \omega_0 - \frac{1}{\sqrt{3}} \omega_0 \right)
    = \mathbf{0}.
\end{align*}
While this demonstrates that our method is not without critical points, even in this simple example, it shows that this configuration is only problematic when the true rotations are equally spaced around an axis of rotation, $\omega_0$, and the estimates are initialized at identity. This compares very favorably to the $SO(3)$ algorithm, which can be in a critical point for any set of relative rotations, $R_i^j$, and with initialization that can vary with an additional arbitrary constant rotation $R_0$.


\section{Method Details}
\label{appendix:method_details}

A full description of the $SO(3)$ Averaging and  Iterative Modified Rodrigues Projective Averaging is shown in Algorithm~\ref{alg:so3_consensus} and Algorithm~\ref{alg:mrp_consensus}, respectively. 
In practice, we find $\gamma=0.5$ and $\eta=0.1$ to produce the best results.

\begin{algorithm}
\SetKwComment{Comment}{//}{}
\SetKwFunction{Update}{Update}
\SetKwInOut{Input}{input}
\SetKwInOut{Output}{output}

\Input{Initial estimates $\hat{\mathcal{R}} = \{\hat{R}_1 \dots \hat{R}_N\}$}
\Input{Local neighborhood $\mathcal{N}_i$ for each rotation $\hat{R}_i$}
\Input{Relative rotations $R_i^j,\; \forall j \in \mathcal{N}_i$}
\Input{Learning Rate $\gamma$}
\Output{Optimized set $\hat{R}_i \in \hat{\mathcal{R}}$}

\caption{$SO(3)$ Averaging}
\label{alg:so3_consensus}
\SetAlgoLined
\While{Not Converged}{
 Sample a rotation $\hat{R}_i \sim \hat{\mathcal{R}}$ to update \\
 Sample a neighbor $\hat{R}_j \sim \mathcal{N}_i$\\
 Compute optimal update $r_\Delta$ with Equation~3b in the main paper \\
 Apply update in $SO(3)$: $\hat{R}_i$ $\gets$ $\hat{R}_i \exp( \gamma r_\Delta)$ \\
}
\Return {$\hat{\mathcal{R}}$}
\end{algorithm}

\begin{algorithm}
\SetKwComment{Comment}{//}{}
\SetKwFunction{Update}{Update}
\SetKwInOut{Input}{input}
\SetKwInOut{Output}{output}

\Input{Initial estimates $\hat{\Psi} = \{\hat{\psi}_1 \dots \hat{\psi}_N\}$}
\Input{Local neighborhood $\mathcal{N}_i$ for each rotations $\hat{\psi}_i$}
\Input{Relative rotations $q_i^j$ for each $j \in \mathcal{N}_i$}
\Input{Learning Rate $\gamma$}
\Input{Max gradient $\eta$}
\Output{Optimized set $\hat{\psi}_i \in \hat{\Psi}$}

\caption{Iterative Modified Rodrigues Projective Averaging}
\label{alg:mrp_consensus}
\SetAlgoLined
\While{Not Converged}{
 Sample a projected rotation $\hat{\psi}_i$ to update \\
 Sample a neighbor $\hat{\psi}_j \sim \mathcal{N}_i$ \\
 Update the projected rotation $\psi_\Delta$ with Equation~5 in the main paper) \\
 \If{the magnitude of the update is larger than $\eta$}{
  Resize update to be of size $\eta$:  $\psi_\Delta \gets \eta \frac{\psi_\Delta}{\|\psi_\Delta\|}$\\
 }
 Apply update in MRP space $\hat{\psi}_i$ $\gets$ {$\hat{\psi}_i + \gamma \psi_\Delta$}
}
\Return {$\hat{\Psi}$}
\end{algorithm}

\section{1DSfM Datasets}
\label{appendix:1dsfm_dataset}

\begin{figure}[h]
 	
    \centering
    \includegraphics[width=\textwidth,trim={0 50 0 0},clip]{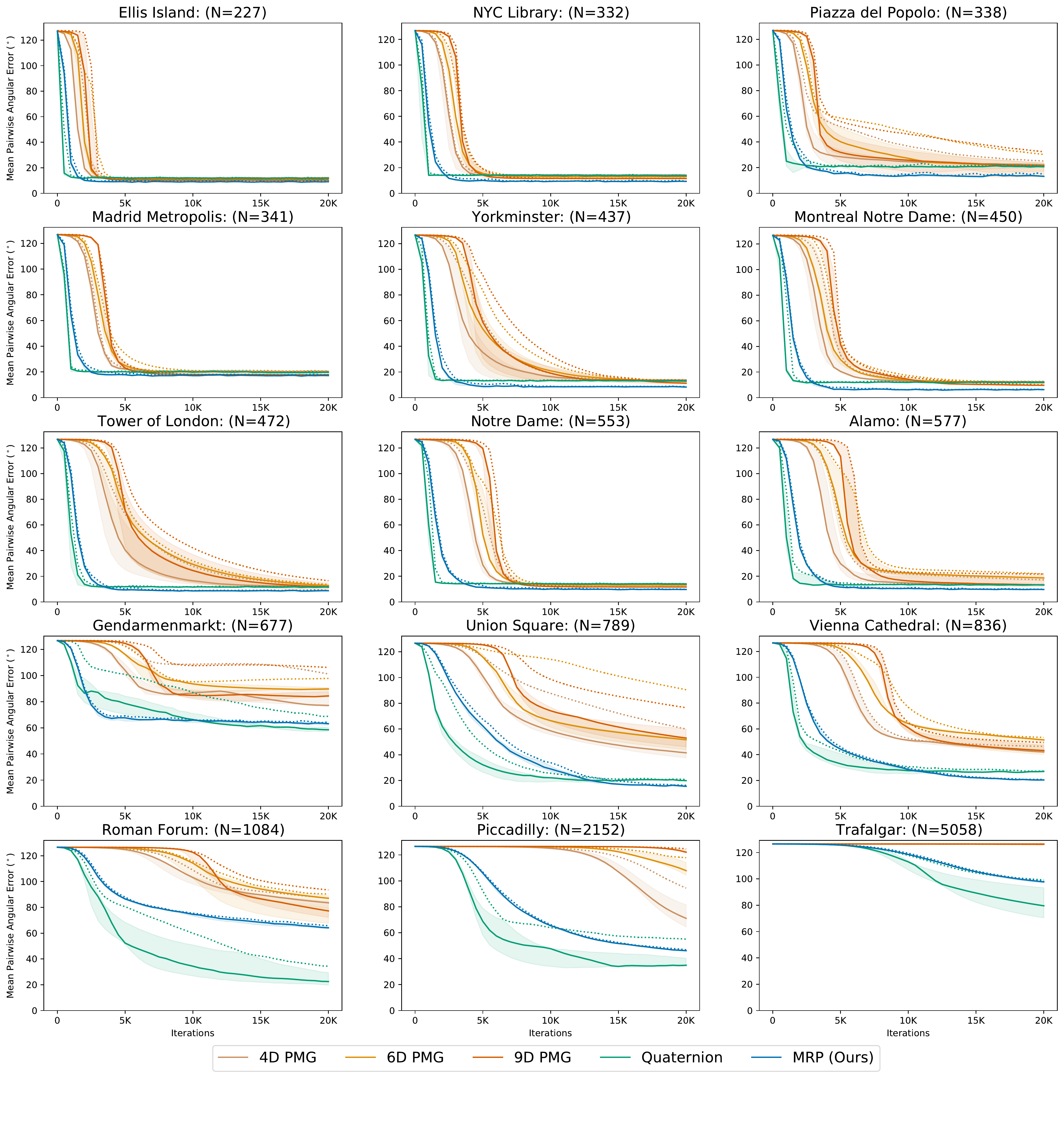}
 	\caption{
 	Optimization results for all 1DSfM~\cite{wilson_eccv2014_1dsfm} datasets, ordered by number of cameras (N). Median average-pairwise angular error (${}^\circ$) is shown with shaded areas representing the first and third quartile over all training sessions. The max average-pairwise angular error for each algorithm at each iteration is shown as a dashed line.}
 	\label{fig:1dsfm_all}
\end{figure}

We report results on all structure from motions datasets available in the 1DSfM~\cite{wilson_eccv2014_1dsfm}. Each environment is tested with 5 random initializations and the estimated rotations are updated by each algorithm in batches of size 64, for 20K iterations. While Iterative Modified Rodrigues Projective Averaging, \textbf{MRP (Ours)} outperform all \textbf{PMG~\cite{chen2021projective}} based methods, the direct \textbf{Quaternion} optimization regularly converges to relatively accurate local optima more quickly than ours, as shown in Table~\ref{tab:all_1dsfm_mean_nauc} and Figure~\ref{fig:1dsfm_all}. That being said, our method converges to a more accurate final configuration for most datasets, with respect to mean relative error, Table~\ref{tab:all_1dsfm_mean_rel}, mean absolute error, Table~\ref{tab:all_1dsfm_mean_abs}, and median absolute error, Table~\ref{tab:all_1dsfm_med_abs}. Our method, as well as the baselines, do not appear to perform well on the larger datasets. As a reminder, this algorithm is specifically designed for training deep learned methods, not for direct rotation optimization. When training deep learned methods, all of the weights are shared, allowing the network to use a single example to improve the accuracy of all rotations near that example. Additionally, we see poor performance on datasets with extremely large observation noise, specifically Gendarmenmarkt, whose median observation error is over 12 degrees. All dataset statistics can be found in Table~\ref{tab:all_1dsfm_stats}. It should be noted that these datasets do not fully cover the orientation space, and tend to largely cover only variations in yaw. For results on datasets that represent full coverage of the orientation space, see the Uniformly Sampled Rotations dataset or the Neural Network Optimization dataset. 

\begin{table}[h]
\scriptsize
    \centering
\begin{tabular}{c|ccccc||ccc}
\hline
\multirow{3}{*}{Dataset} & \multicolumn{8}{c}{\textit{Mean Absolute Error (${}^{\circ}$)}}\\
\hline
 &   4D PGM &  6D PGM &  9D PGM &  Quat &   MRP (Ours) &  IRLS-GM &  IRLS-$\ell_{1/2}$ & MLP \\
 &~\cite{chen2021projective} &~\cite{ zhou2019continuity} & ~\cite{levinson2020analysis}& & & \cite{chatterjee2013efficient} & \cite{chatterjee2017robust} & \cite{shi2020message}\\
\hline
 Ellis Island        &     7.5  &     7.03 &     6.41 &   7.44 &         \textbf{5.59} & 3.04 & 2.71 & 2.61 \\
 NYC Library         &     9.23 &     8.32 &     7.38 &   8.92 &         \textbf{6.03} & 2.71 & 2.66 & 2.63\\
 Piazza del Popolo   &    16.37 &    16.1  &    15.88 &  15.24 &        \textbf{10.03} & 4.10 & 3.99 & 3.73\\
 Madrid Metropolis   &    13.55 &    13.23 &    11.78 &  13    &        \textbf{11.25} & 5.30 & 4.88 & 4.65\\
 Yorkminster         &     9.13 &     8.34 &     7.48 &   8.56 &         \textbf{5.3} & 2.60 & 2.45 & 2.47\\
 Montreal Notre Dame &     8.17 &     7.65 &     6.24 &   7.76 &         \textbf{4.02} & 2.63 & 2.26 & 2.06\\
 Tower of London     &     8.02 &     8.12 &     8.36 &   7.44 &         \textbf{5.58} & 3.42 & 3.41 & 3.16\\
 Notre Dame          &     8.71 &     7.96 &     7.03 &   8.55 &         \textbf{5.80} & 2.63 & 2.26 & 2.06 \\
 Alamo               &     9.41 &    11.98 &    10.98 &   8.74 &         \textbf{6.42} & 3.64 & 3.67 & 3.44\\
 Gendarmenmarkt      &    66.41 &    73.7  &    68.29 &  \textbf{46.63} &        48.82 & 39.24 & 39.41 & 44.94\\
 Union Square        &    32.46 &    40.86 &    40.92 &  13.44 &        \textbf{10.22} & 6.77 & 6.77 & 6.54\\
 Vienna Cathedral    &    29.18 &    31.42 &    32.94 &  18.67 &        \textbf{13.60} & 8.13 & 8.07 & 7.21\\
 Roman Forum         &    63.23 &    64.85 &    60.51 &  \textbf{18.11} &        55.65 & 2.66 & 2.69 & 2.62\\
 Piccadilly          &    53.35 &    84.37 &   106.84 &  \textbf{26.29} &        29.98 & 5.12 & 5.19 & 3.93\\
 Trafalgar           &   121.93 &   124.18 &   125.15 &  \textbf{69.65} &        91.67 & - & - & - \\
\hline
\end{tabular}
\caption{\textbf{Final Mean Absolute Rotation Error Results on 1DSfM ~\cite{wilson_eccv2014_1dsfm} dataset.} Results on the left before the double lines are comparisons of local method after 20K iterations. Results on the right after the double lines are obtained from global methods which require optimizing over global set of relative orientations data at each step. Results associated sections with dashed line are not available from global methods~\cite{shi2020message}.}
    \label{tab:all_1dsfm_mean_abs}
\end{table}

\begin{table}[h]
\scriptsize
    \centering
\begin{tabular}{c|ccccc||ccc}
\hline
\multirow{3}{*}{Dataset} & \multicolumn{8}{c}{\textit{Median Absolute Error (${}^{\circ}$)}}\\
\hline
 &  4D PGM &  6D PGM & 9D PGM & Quat & MRP (Ours) &  IRLS-GM &  IRLS-$\ell_{1/2}$ & MLP \\
 &~\cite{chen2021projective} &~\cite{ zhou2019continuity} & ~\cite{levinson2020analysis}& & & \cite{chatterjee2013efficient} & \cite{chatterjee2017robust} & \cite{shi2020message}\\
\hline
 Ellis Island        &     3.68 &     3.25 &     3.12 &   4.04 &         \textbf{2.96} & 1.06 & 0.93 & 0.88\\
 NYC Library         &     6.11 &     5.52 &     4.85 &   6.11 &         \textbf{4.04} & 1.37 & 1.30 & 1.24\\
 Piazza del Popolo   &     9.51 &     9.32 &     9.32 &   9.29 &         \textbf{6.12} & 2.17 & 2.09 & 1.93\\
 Madrid Metropolis   &     9.37 &     9.06 &     7.86 &   9.07 &         \textbf{6.99} & 1.78 & 1.88 & 1.26\\
 Yorkminster         &     6.44 &     5.77 &     4.56 &   6.11 &         \textbf{3.29} & 1.59 & 1.53 & 1.45\\
 Montreal Notre Dame &     3.86 &     3.56 &     2.86 &   3.90  &         \textbf{2.30} & 0.58 & 0.57 & 0.51\\
 Tower of London     &     4.87 &     5.84 &     6.36 &   4.64 &         \textbf{3.59} & 2.52 & 2.50 & 2.20\\
 Notre Dame          &     4.39 &     3.73 &     3.09 &   4.48 &         \textbf{2.61} & 0.78 & 0.71 & 0.67\\
 Alamo               &     4.73 &     5.77 &     5.16 &   4.90  &         \textbf{3.48} & 1.30 & 1.32 & 1.16\\
 Gendarmenmarkt      &    64.08 &    71.57 &    62.9  &  \textbf{43.91} &        45.92 & 7.07 & 7.12 & 9.87\\
 Union Square        &    27.75 &    34.68 &    34.84 &   9.75 &         \textbf{6.85} & 3.66 & 3.85 & 3.48\\
 Vienna Cathedral    &    13.80 &    13.77 &    16.73 &  11.67 &         \textbf{6.34} & 1.92 & 1.76 & 2.83\\
 Roman Forum         &    53.78 &    62.46 &    57.71 &  \textbf{16.56} &        41.95 & 1.58 & 1.57 & 1.37\\
 Piccadilly          &    42.34 &    79.74 &   107.32 &  19.67 &        \textbf{15.09} & 2.02 & 2.34 & 1.81\\
 Trafalgar           &   126.71 &   129.57 &   130.45 &  \textbf{65.54} &        89.09 & - & - & -\\
\hline
\end{tabular}
    \caption{\textbf{Final Median Absolute Rotation Error Results on 1DSfM ~\cite{wilson_eccv2014_1dsfm} dataset.} Results on the left before the double lines are comparisons of local method after 20K iterations. Results on the right after the double lines are obtained from global methods which require optimizing over global set of relative orientations data at each step. Results associated sections with dashed line are not available from global methods~\cite{shi2020message}.}
    \label{tab:all_1dsfm_med_abs}
\end{table}

\begin{table}[h]
\scriptsize
    \centering
\begin{tabular}{c|ccccc}
\hline
\multirow{2}{*}{Dataset} & \multicolumn{5}{c}{\textit{Mean nAUC}} \\
 &  4D PGM~\cite{chen2021projective} &  6D PGM~\cite{ zhou2019continuity}   &   9D PGM~\cite{ levinson2020analysis} &   Quat &   MRP (Ours) \\
\hline
 Ellis Island        &    22.56 &    24.07 &    25.02 &  15.05 &        \textbf{14.58} \\
 NYC Library         &    28.53 &    31.12 &    32.07 &  18.20  &        \textbf{16.84} \\
 Piazza del Popolo   &    37.36 &    44.18 &    43.98 &  25.13 &        \textbf{22.21} \\
 Madrid Metropolis   &    35.91 &    38.49 &    39.15 &  \textbf{24.34} &        24.48 \\
 Yorkminster         &    36.82 &    42.37 &    44.91 &  18.71 &        \textbf{18.43} \\
 Montreal Notre Dame &    33.97 &    37.54 &    40.37 &  17.69 &        \textbf{16.19} \\
 Tower of London     &    39.98 &    45.99 &    49.54 &  \textbf{18.14} &        18.85 \\
 Notre Dame          &    38.77 &    43.04 &    46.05 &  \textbf{20.78} &        21.10  \\
 Alamo               &    39.87 &    49.08 &    50.22 &  \textbf{20.47} &        22.05 \\
 Gendarmenmarkt      &    97.45 &   101.77 &   100.11 &  74.76 &        \textbf{71.39} \\
 Union Square        &    77.22 &    87.01 &    89.76 &  \textbf{34.60}  &        46.20  \\
 Vienna Cathedral    &    72.25 &    81.07 &    83.48 &  \textbf{38.74} &        42.94 \\
 Roman Forum         &   103.59 &   105.73 &   108.88 &  \textbf{52.05} &        82.30  \\
 Piccadilly          &   115.83 &   123.41 &   126.16 &  \textbf{62.87} &        78.31 \\
 Trafalgar           &   126.43 &   126.49 &   126.5  & \textbf{108.19} &       115.90  \\
\hline
\end{tabular}
    \caption{Final Mean Normalized AUC on all 1DSfM~\cite{wilson_eccv2014_1dsfm} datasets after 20K iterations}
    \label{tab:all_1dsfm_mean_nauc}
\end{table}
\begin{table}[h]
\scriptsize
    \centering
\begin{tabular}{c|ccccc}
\hline
\multirow{2}{*}{Dataset}  & \multicolumn{5}{c}{\textit{Mean Relative Error (${}^{\circ}$)}}\\
 &  4D PGM~\cite{chen2021projective} & 6D PGM~\cite{ zhou2019continuity}   &  9D PGM~\cite{ levinson2020analysis} &   Quat &   MRP (Ours) \\
\hline
 Ellis Island        &    12.21 &    11.49 &    10.37 &  11.87 &         \textbf{9.03} \\
 NYC Library         &    14.29 &    12.94 &    11.51 &  13.67 &         \textbf{9.30}  \\
 Piazza del Popolo   &    21.91 &    21.24 &    20.64 &  20.74 &        \textbf{13.49} \\
 Madrid Metropolis   &    20.43 &    19.84 &    17.85 &  19.62 &        \textbf{17.09} \\
 Yorkminster         &    13.73 &    12.64 &    11.58 &  12.97 &         \textbf{8.35} \\
 Montreal Notre Dame &    12.5  &    11.59 &     9.58 &  11.93 &        \textbf{6.22} \\
 Tower of London     &    12.41 &    12.24 &    12.44 &  11.56 &         \textbf{8.71} \\
 Notre Dame          &    14.15 &    13.1  &    11.65 &  13.86 &         \textbf{9.66} \\
 Alamo               &    14.23 &    17.47 &    15.75 &  13.17 &         \textbf{9.78} \\
 Gendarmenmarkt      &    84.21 &    89.61 &    84.77 &  \textbf{60.25} &        62.98 \\
 Union Square        &    44.44 &    55.4  &    55.94 &  19.98 &        \textbf{15.52} \\
 Vienna Cathedral    &    41.8  &    45.62 &    44.18 &  26.64 &        \textbf{20.32} \\
 Roman Forum         &    79.24 &    77.18 &    78.03 &  \textbf{25.04} &        64.25 \\
 Piccadilly          &    74.25 &   105.15 &   122.06 &  \textbf{38.61} &        46.21 \\
 Trafalgar           &   126.18 &   126.42 &   126.49 &  \textbf{81.28} &        97.53 \\
\hline
\end{tabular}
    \caption{Final Mean Relative Error $\left( {}^\circ \right)$  on all 1DSfM~\cite{wilson_eccv2014_1dsfm} datasets after 20K iterations}
    \label{tab:all_1dsfm_mean_rel}
\end{table}
\begin{table}[h]
\scriptsize
    \centering
\begin{tabular}{c|cccc}
\hline
 Dataset             &   \# Nodes &   \# Edges &   Mean Error &   Median Error \\
\hline
\hline
 Ellis Island        &       227 & 20K       &        12.52 &        2.89 \\
 NYC Library         &       332 & 21K       &        14.15 &        4.22 \\
 Piazza del Popolo   &       338 & 25K       &         8.4  &        1.81 \\
 Madrid Metropolis   &       341 & 24K       &        29.31 &        9.34 \\
 Yorkminster         &       437 & 28K       &        11.17 &        2.68 \\
 Montreal Notre Dame &       450 & 52K       &         7.54 &        1.67 \\
 Tower of London     &       472 & 24K       &        11.6  &        2.59 \\
 Notre Dame          &       553 & 104K      &        14.16 &        2.7  \\
 Alamo               &       577 & 97K       &         9.1  &        2.78 \\
 Gendarmenmarkt      &       677 & 48K       &        33.33 &       12.3  \\
 Union Square        &       789 & 25K       &         9.03 &        3.61 \\
 Vienna Cathedral    &       836 & 103K      &        11.28 &        2.59 \\
 Roman Forum         &      1084 & 70K       &        13.84 &        2.97 \\
 Piccadilly          &      2152 & 309K      &        19.1  &        4.93 \\
 Trafalgar           &      5058 & 679K      &         8.64 &        3.01 \\
\hline
\end{tabular}
    \caption{Dataset sizes and observation accuracies $\left( {}^\circ \right)$ for  all 1DSfM~\cite{wilson_eccv2014_1dsfm} datasets}
    \label{tab:all_1dsfm_stats}
\end{table}

\section{Curriculum for Neural Network Optimization}
\label{appendix:curriculum}

\begin{figure}{h}
 	\centering
 	\vspace{-1em}

     	\includegraphics[width=0.6\textwidth, trim={10 0 40 15}, clip]{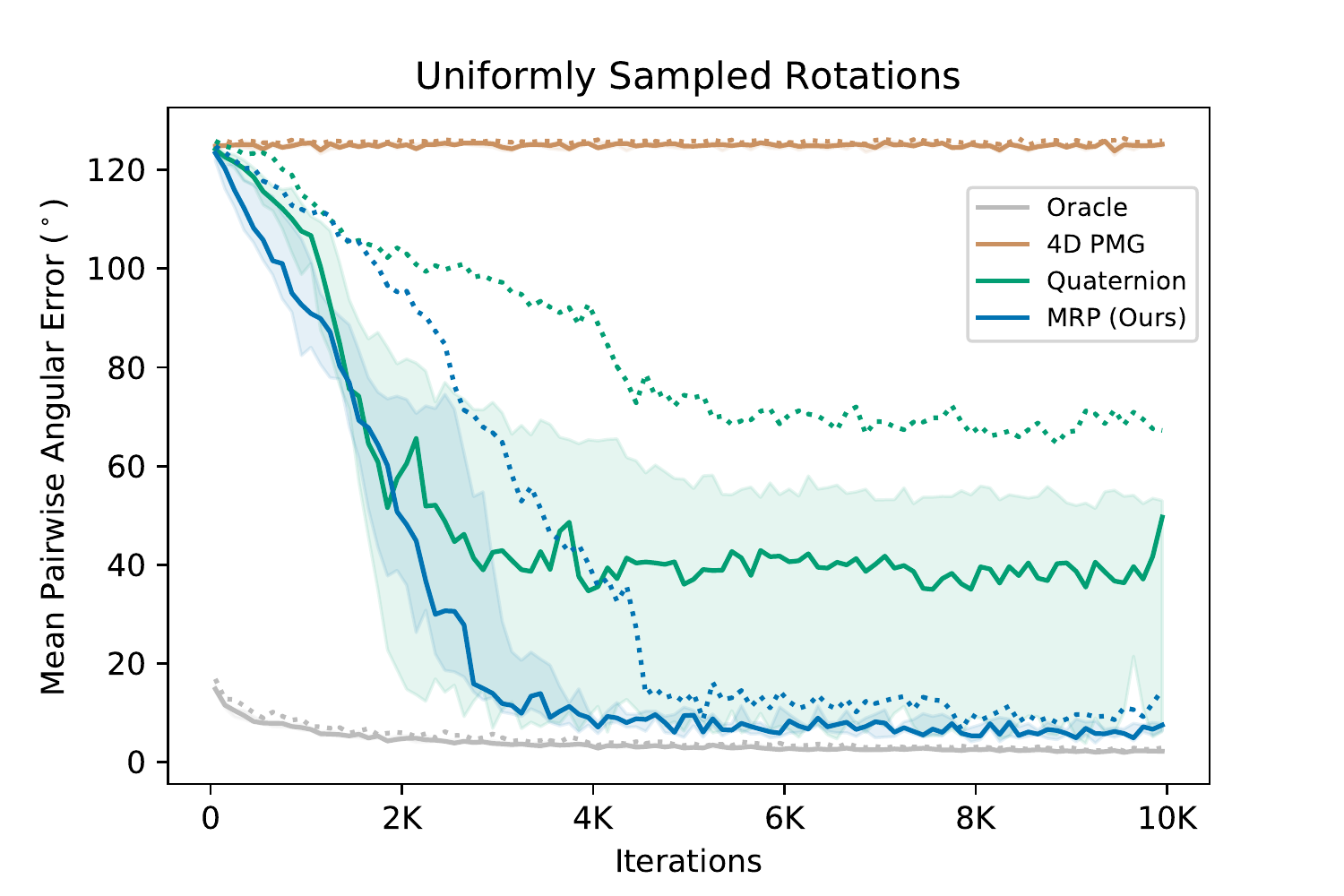}

 	\captionof{figure}{
 	Training results for rotations estimated by neural networks given images of the YCB drill~\cite{xiang2018posecnn} rendered at each of 100 random rotations with various supervisions.
    Median average-pairwise angular error (${}^\circ$) is shown with shaded areas representing the first and third quartile over all training sessions. The max average-pairwise angular error for each algorithm at each iteration is shown as a dashed line.
    }
 	\label{fig:nn_convergence}
\end{figure}

\begin{figure}[ht]
 	
    \centering
    \includegraphics[width=.49\textwidth,trim={0 0 0 0},clip]{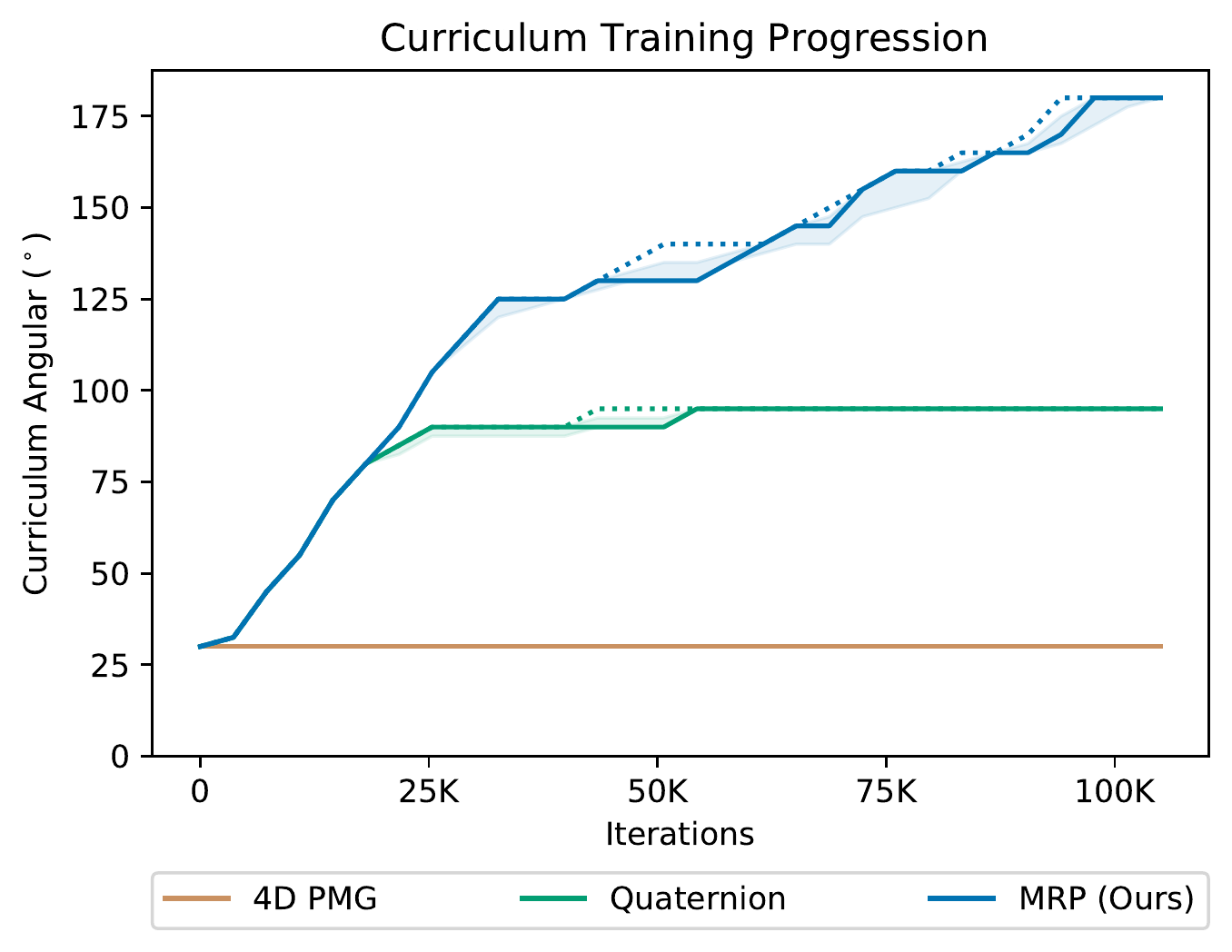}
    \hfill    \includegraphics[width=.49\textwidth,trim={0 0 0 0},clip]{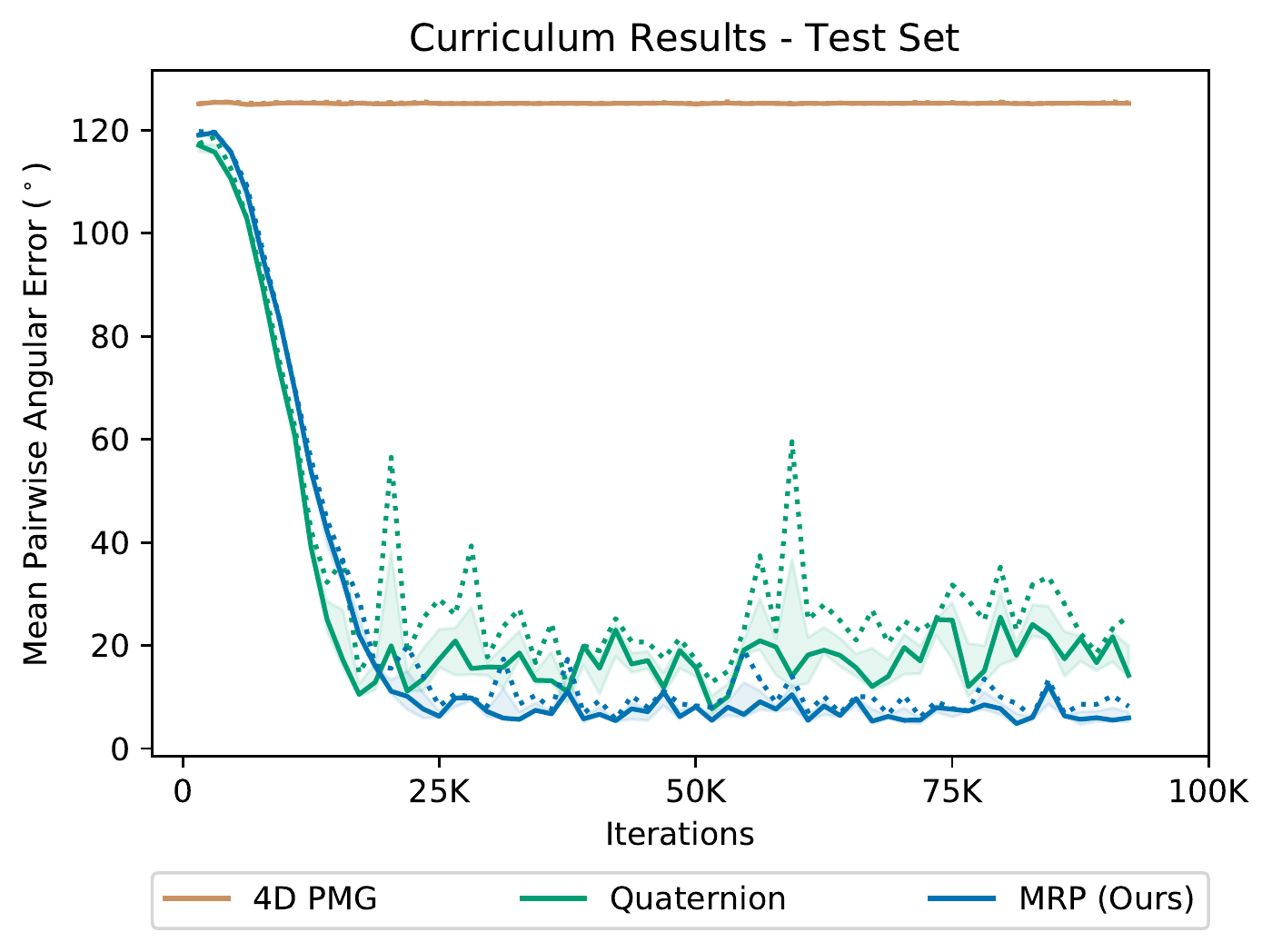}   	
 	\caption{Curriculum Angle (left) and Average Pairwise Error (right), sampled over the full orientation space for three training sessions with each method. Median average-pairwise angular error (${}^\circ$) is shown with shaded areas representing the first and third quartile over all training sessions. The max average-pairwise angular error for each algorithm at each iteration is shown as a dashed line.}
\label{fig:curr_results}
\end{figure}

While the \textbf{MRP (Ours)} was able to learn the orientations of a fixed set of rotation images, training results shown in Figure~\ref{fig:nn_convergence}, we find that a curriculum is required for any relatively supervised method to generalized to unseen orientation. This  curriculum training involves starting with a initial base rotation. The model is rendered at this base rotation and a random rotation within $30^\circ$ of this base rotation. This base rotation is initially sampled with $\theta=30^\circ$ of a constant anchor orientation, until the average training angular error of the previous epoch drops below a given threshold, in this case, $5^\circ$. Once the error drops below this threshold, the angular range, $\theta$, from which this base rotation is sampled is increased by $5^\circ$. This process is repeated, increasing the value of $\theta$ by $5^\circ$ each time the error threshold is reached. We find that \textbf{MRP (Ours)} is able to complete the curriculum in a reasonable number of iterations, about 100K, achieving a median final pairwise accuracy of $5.19^\circ$ over three training sessions. This test error is sampled from two random rotations across the $SO(3)$, differing from the training error, which are sampled based on the curriculum and are always, at most, $30^\circ$ apart. The quaternion optimization method, \textbf{Quaternion}, stalls out at curriculum angle of $90^\circ$, achieving a final pairwise accuracy of $12.41^\circ$ and the   \textbf{4D PMG~\cite{chen2021projective}} method never gets past the first level of the curriculum, with a final error of $125.09^\circ$.
The full training progression of each method, over three random initialization each, can be seen in Figure~\ref{fig:curr_results}

One way this curriculum could be applied to captured data as follows: given a video, a curriculum could be established based on temporal proximity in the video. Choosing an arbitrary initial frame of the video as a anchoring frame, a curriculum can be generate by increasing temporal distance to neighboring frames until the entire video has been used in training.

\section{3D Object Rotation Estimation via Relative Supervision from Pascal3D+ Images}
\label{appendix:pascal_exp}
\subsection{Experimental Setup} 
\label{appendix:pascal_exp_settings}
Pascal3D+~\cite{xiang2014beyond} is a standard benchmark for categorical 6D object pose estimation from real images. We follow similar experimental settings as in ~\cite{chen2021projective, levinson2020analysis} for 3D object pose estimation from single images. Following~\cite{chen2021projective, levinson2020analysis}, we discard occluded or truncated objects and augment with rendered images from~\cite{su2015render}. We report 3D object pose estimation via relative orientation supervision results on two object categories of Pascal3D+ image dataset: \textit{sofa} and \textit{bicycle}. We compare our method \textbf{MRP} with five baselines: \textbf{Quaternion}, \textbf{4D PMG}~\cite{chen2021projective}, \textbf{6D PMG} ~\cite{ zhou2019continuity}, \textbf{9D PMG} ~\cite{ levinson2020analysis} and \textbf{10D PMG}~\cite{ peretroukhin2020smooth}.

We use ResNet18~\cite{he2016deep} as the model backbone to predict object rotation from single images. The model is supervised by the geodesic error between the induced relative orientation between the predicted absolute orientations for a pair of images, and the relative orientation between the ground truth absolute orientations for the image pair. 

Specifically, \textbf{MRP} is supervised by the geodesic distance on the MRP manifold as described in Equations~4 and~5 in the main paper. \textbf{Quaternion} is supervised by quaternion geodesic distance as described in Section 4 in the main paper. While \textbf{4D/6D/9D/10D PMG} are supervised by the geodesic error derived from projective manifold gradients as in ~\cite{chen2021projective}. 
We use the same batch size of 20 as in~\cite{chen2021projective, levinson2020analysis}, and use Adam~\cite{kingma2015adam} with learning rate of 1e-4.

\subsection{Result Analysis}
\begin{figure}[b]
 	\centering
     	\includegraphics[width=0.6\textwidth]{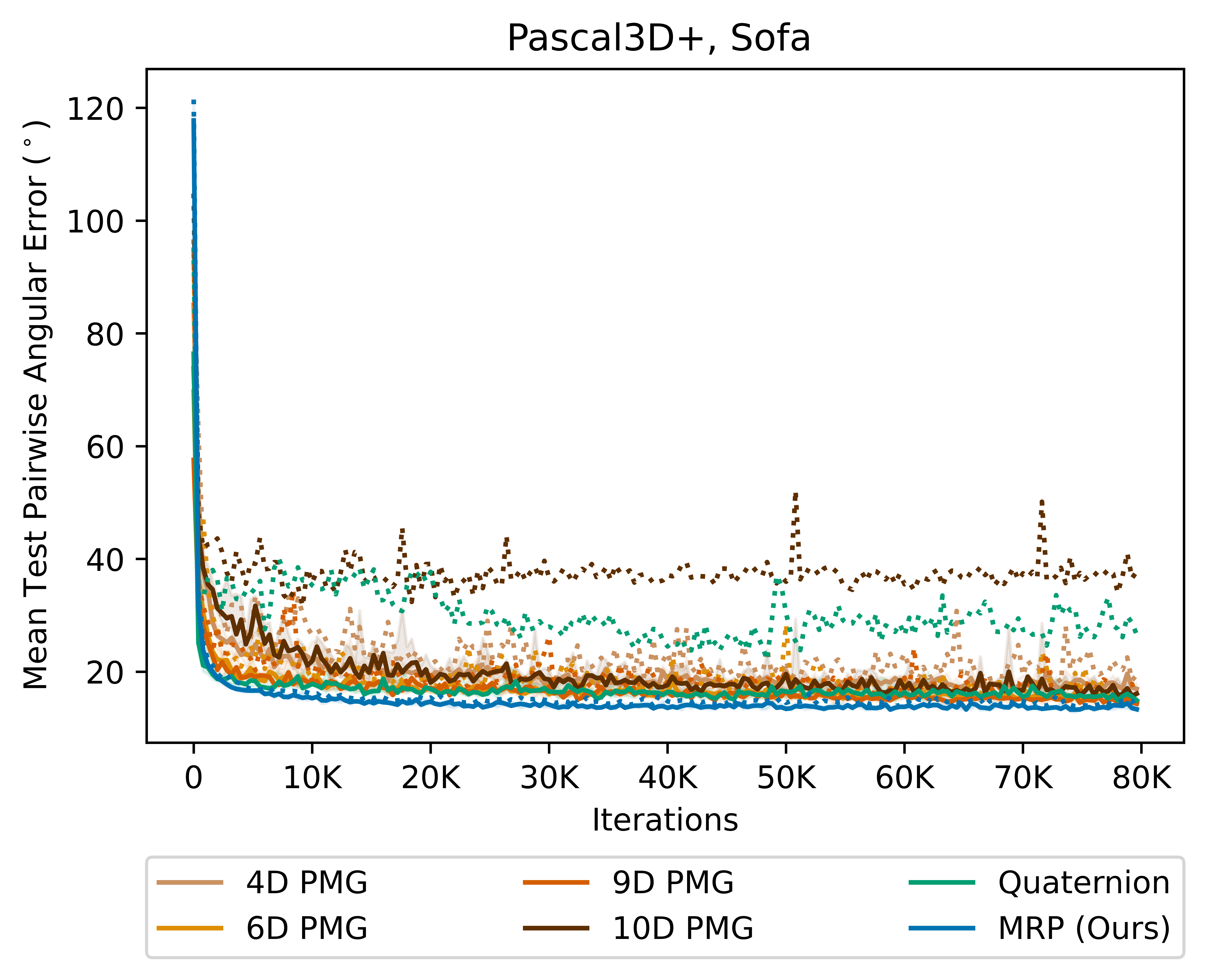}
\caption{\textbf{3D Object Pose Estimation via Relative Supervision on Pascal3D+ \textit{Sofa} Images.} Mean test pairwise angular error in degrees of \textit{sofa} at different iterations of training. Trained over 80K training steps for 8 random seeds per method. Solid lines stand for mean errors, dashed line stand for max errors, and shaded area represents error standard deviation.} 
\label{fig:pascal_sofa}
\end{figure}

\begin{figure}[h]
 	\centering
     	\includegraphics[width=0.6\textwidth]{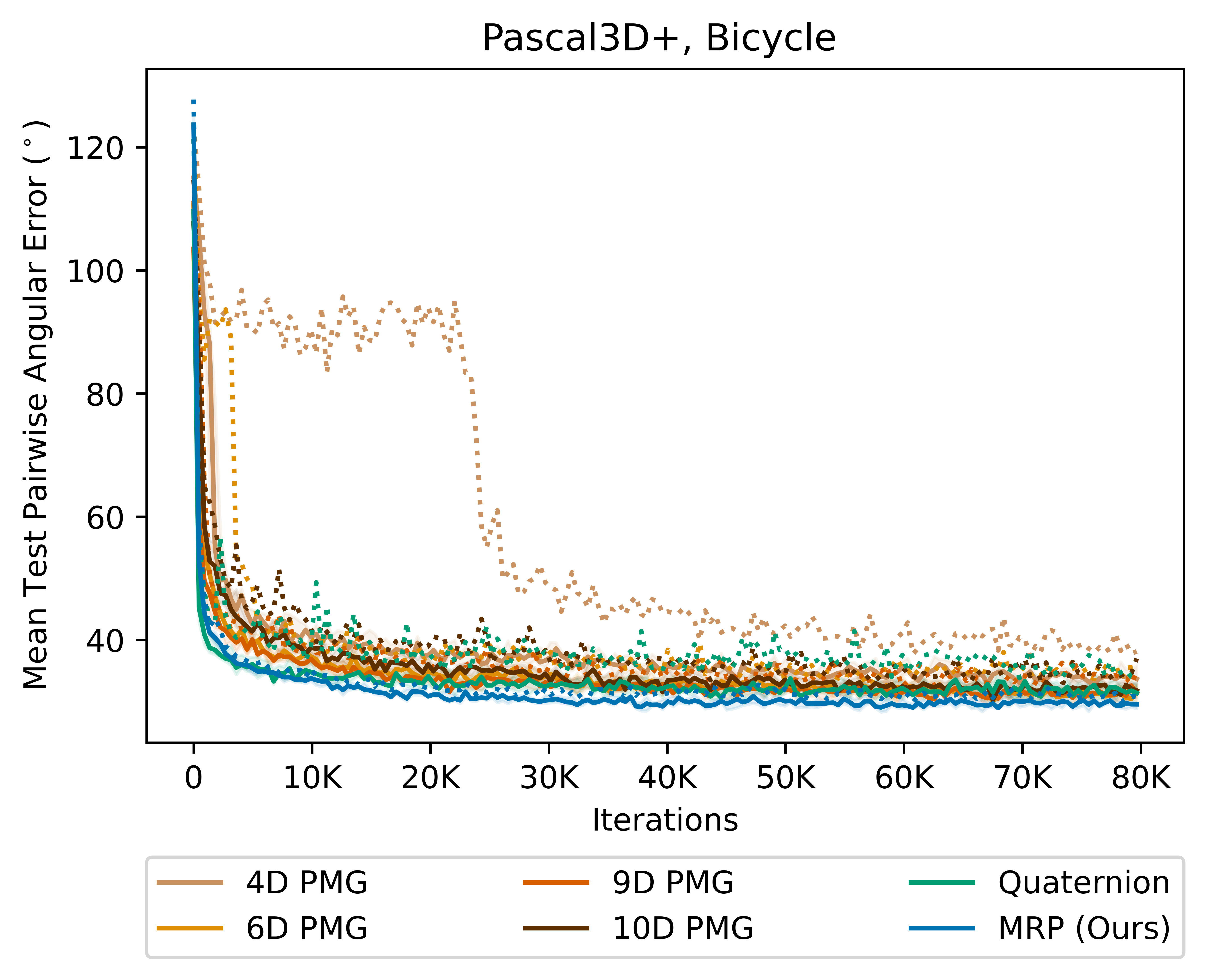}
     	
\caption{\textbf{3D Object Pose Estimation via Relative Supervision on Pascal3D+ \textit{Bicycle} Images.} Mean test pairwise angular error (${}^\circ$) of \textit{bicycle} at different iterations of training. Trained over 80K training steps for 8 random seeds per method. Solid lines stand for mean errors, dashed line stand for max errors, and shaded area represents error standard deviation.
 	 } 
 
\label{fig:pascal_bike}
 	 
\end{figure}

Results for \textit{sofa} showed in Figure~\ref{fig:pascal_sofa} and Table~\ref{tab:pascal_sofa}. Results for \textit{bicycle} showed in Figure~\ref{fig:pascal_bike} and Table~\ref{tab:pascal_bike}. 
\textbf{Pascal3D+ Sofa.} For \textit{sofa} category, as seen in Table~\ref{tab:pascal_sofa}, we find that after 50K training iterations, \textbf{MRP (Ours)} achieves a mean angular pairwise error of $14.09^{\circ}$ on the test set, outperforms all other baselines. \textbf{Quaternion} achieves the worst error out of all methods, with final angular pairwise error of $26.35^{\circ}$. Besides achieving the lowest test angular error, we also find that $\textbf{MRP (Ours)}$ has the fastest convergence speed, as seen in Figure~\ref{fig:pascal_sofa}.
\begin{table}[b]
    \centering
    \begin{tabular}{c|c c c}
    \hline
     Algorithm  &  Mean (${}^{\circ}$) & Min (${}^{\circ}$) & Max (${}^{\circ}$)\\
    \hline
     4D PMG~\cite{chen2021projective} & 17.39 $\pm$ 1.14 & 19.42 & 16.07\\
     6D PMG~\cite{zhou2019continuity} &  15.20 $\pm$ 0.77 & 16.43 & 14.44 \\
     9D PMG~\cite{levinson2020analysis} &  14.61 $\pm$ 0.50 & 15.66 & 14.18\\
     10D PMG~\cite{peretroukhin2020smooth} & 19.28 $\pm$ 7.58 & 37.76 & 15.03\\ 
     Quaternion &  16.52 $\pm$ 4.12 & 26.57 & 14.38\\ 
     \textbf{MRP (Ours)} &  \textbf{13.63 $\pm$ 0.78} & \textbf{15.08} & \textbf{12.62}\\
    \hline
    \end{tabular}
    \caption{\textbf{Final Mean Test Angular Pairwise Error on Pascal3D+ \textit{sofa} Images after 80K training iterations, over 8 random seeds.}}
    \label{tab:pascal_sofa}
\end{table}

\textbf{Pascal3D+ Bicycle.} For \textit{bicycle} category, as seen in Table~\ref{tab:pascal_bike}, we find that after 50K training iterations, \textbf{MRP (Ours)} achieves a mean angular pairwise error of $29.21^{\circ}$ on the test set, outperforms all other baselines. Besides achieving the lowest test angular error, we also find that $\textbf{MRP (Ours)}$ has the fastest convergence speed, as seen in Figure~\ref{fig:pascal_bike}.
\begin{table}[H]
    \centering
    \begin{tabular}{c|ccc}
    \hline    
     Algorithm  &  Mean (${}^{\circ}$) & Min (${}^{\circ}$) & Max (${}^{\circ}$) \\
    \hline
     4D PMG~\cite{chen2021projective} & 34.57 $\pm$ 2.21 & 38.13 & 31.90 \\
     6D PMG~\cite{zhou2019continuity} &  31.58 $\pm$ 2.24 & 35.66 & \textbf{28.42}\\
     9D PMG~\cite{levinson2020analysis} &  31.80 $\pm$ 1.52 & 34.87 & 29.96  \\
     10D PMG~\cite{peretroukhin2020smooth} &  32.23 $\pm$ 2.10 & 36.98 & 29.87 \\ 
     Quaternion &  31.92 $\pm$ 1.00 & 33.61 & 30.61\\ 
     \textbf{MRP (Ours)} &  \textbf{29.46 $\pm$ 0.66} & \textbf{30.74} & 28.62\\

    \hline
    \end{tabular}
    \caption{\textbf{Final Mean Test Angular Pairwise Error on Pascal3D+ \textit{bicycle} Images after 80K training iterations, over 8 random seeds.}}
    \label{tab:pascal_bike}
\end{table}

\section{3D Object Rotation Estimation via Relative Supervision from ModelNet40 Point Clouds}
\label{appendix:modelnet40}

\subsection{Experimental Setup} 

ModelNet40~\cite{wu20153d} is a standard benchmark for categorical 6D object pose estimation from 3D point clouds. We follow similar experimental settings as in ~\cite{chen2021projective}. We follow the same train/test data split as in~\cite{chen2021projective} and report 3D object pose estimation via relative orientation supervision results on the \textit{airplane} category of ModelNet40 dataset. We compare our method \textbf{MRP} with four baselines: \textbf{Quaternion}, \textbf{4D PMG}~\cite{chen2021projective}, \textbf{6D PMG} ~\cite{ zhou2019continuity},  \textbf{9D PMG}~\cite{ levinson2020analysis} and \textbf{10D PMG}~\cite{ peretroukhin2020smooth}
.
We use PointNet++~\cite{qi2017pointnet++} as the model backbone to predict 3D absolute object rotation from single point cloud generated from the ModelNet40 3D CAD models, as in ~\cite{chen2021projective}. The model is supervised by the geodesic error between the induced relative orientation between the predicted absolute orientations for a pair of point clouds, and the relative orientation between the ground truth absolute orientations for the point cloud pair.

We sample 1024 points per point cloud as in~\cite{chen2021projective, levinson2020analysis}, use a batch size of 14. As for training, we use Adam~\cite{kingma2015adam} with learning rate of 1e-3, and run over 1 trial for each method.

We find that for any of the compared methods to generalize to unseen test point cloud instances, a curriculum is needed. We train with a curriculum over the rotation space, the curriculum details can be found in Section~\ref{appendix:curriculum}. Specifically we start with base rotation range with $\theta=30^{\circ}$ of a constant anchor orientation, and $\theta$ is increased by $5^{\circ}$ whenever the previous mean epoch train angular error drops below the curriculum threshold, $5^{\circ}$. To speed up the training procedure, we increase this curriculum threshold to $8^{\circ}$ once $\theta$ gets to $125^{\circ}$. 

\subsection{Result Analysis}
\begin{figure}[ht]
 	
    \centering
    \includegraphics[width=.49\textwidth,trim={0 0 0 0},clip]{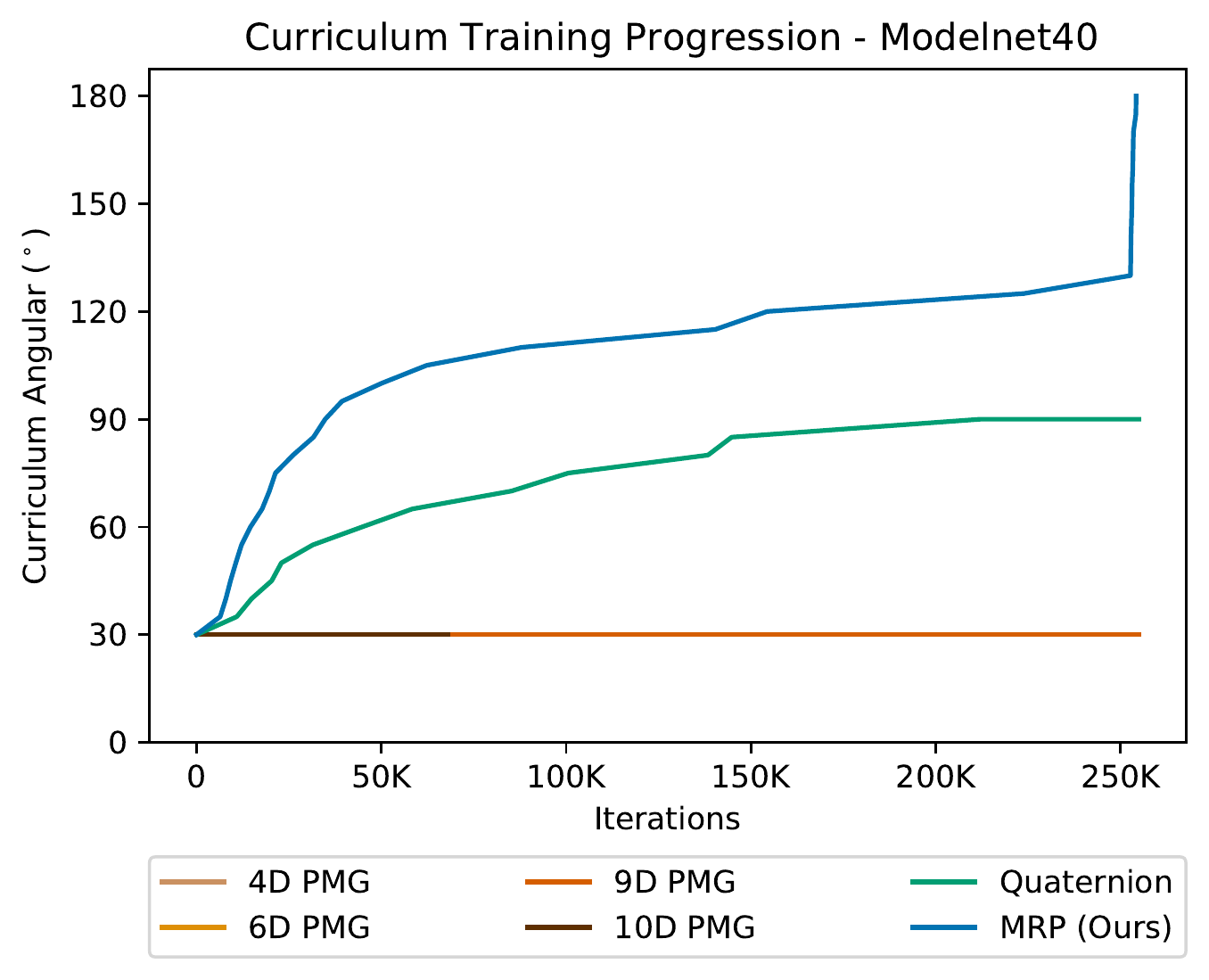}
    \hfill    \includegraphics[width=.49\textwidth,trim={0 0 0 0},clip]{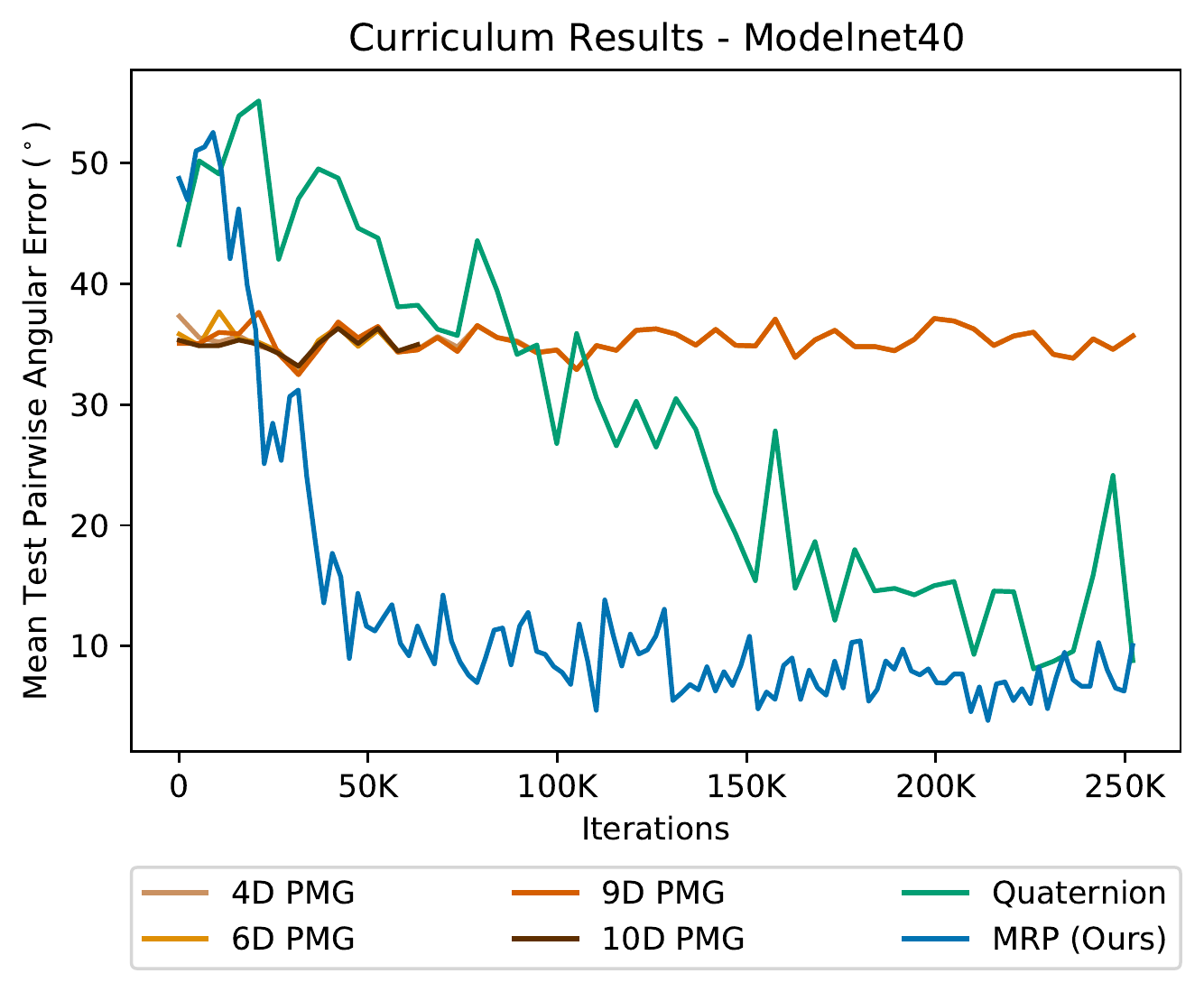}   	
 	\caption{\textbf{3D Object Rotation Estimation via Relative Supervision from ModelNet40 Point Clouds -  \textit{airplane}.}
 	\textbf{Left:} Curriculum angle progression through training iterations. \textbf{Right:} Average test pairwise angular error (${}^{\circ}$), sampled over the full orientation space for 1 training session with each method.}
 	\label{fig:modelnet_curr}
\end{figure}

Results on the \textit{airplane} object class from ModelNet40 dataset is shown in Figure~\ref{fig:modelnet_curr} and Table~\ref{tab:modelnet_airplane}. 

As seen in Figure~\ref{fig:modelnet_curr} and Table~\ref{tab:modelnet_airplane}, \textbf{MRP (Ours)} is able to go through the curriculum in 250K iterations, reaching final test pairwise angular error of $5.49^{\circ}$. \textbf{Quaternion} goes through the curriculum much slower, reaching curriculum angle $\theta = 90^{\circ}$ at the end of 250K steps. \textbf{4D PMG}, \textbf{6D PMG}, \textbf{9D PMG} and \textbf{10D PMG}, on the other hand, is not able to progress beyond the original curriculum angle of $\theta = 30^{\circ}$, reaching final test pairwise angular error around $35^{\circ}$ after 200K iterations. In summary, \textbf{MRP (Ours)} achieves faster convergence rate than all baselines, and is able to achieve final test angular error on the order of $5^{\circ}$ after progressing through the curriculum.

\begin{table}[H]
    \centering
    \begin{tabular}{c|c }
    \hline
     Algorithm  &  Mean Test Angular Pairwise Error (${}^{\circ}$)\\
    \hline
     4D PMG~\cite{chen2021projective} & 35.35 \\
     6D PMG~\cite{zhou2019continuity} &  34.12 \\
     9D PMG~\cite{levinson2020analysis} &  35.80 \\
     10D PMG~\cite{peretroukhin2020smooth} &  35.26 \\
     Quaternion &  12.86\\ 
     \textbf{MRP (Ours)} &  \textbf{5.49}\\
    \hline
    \end{tabular}
    \caption{\textbf{Final Mean Test Angular Pairwise Error on ModelNet40 \textit{airplane} Point Clouds after at most 250K training iterations.}}
    \label{tab:modelnet_airplane}
\end{table}

\section{Absolute Orientation Supervision}
\label{appendix:absolute_supervision}
\subsection{Experimental Setup}
In this paper, we are assuming that only relative orientation supervision is available; however, in this section we explore how different orientation representations perform if absolute orientation supervision is available, and specifically how Modified Rodriguez Parameters (MRP)~\cite{crassidis1996attitude} used in this paper compare.
To explore this, we perform an experiment on rotation estimation from 2D images of rendered YCB drill supervised with absolute orientation instead of relative supervision. We follow the same experimental setup as in Section 6.2 in the main paper, utilizing ResNet18~\cite{he2016deep} as the model backbone to predict absolute 3D object orientations from sets of 2D rendered object images, rendered at 100 random rotations each. The neural network model is supervised by the geodesic error between the predicted absolute orientation and the ground truth absolute orientation. We compare the performance of different rotation parameterizations on this task. Specifically, we compare the Modified Rodriguez Parameters (MRP)~\cite{crassidis1996attitude} (\textbf{Oracle-MRP}) with Quaternions (\textbf{Oracle-Quaternion}). Each method is trained for 10K steps, over 8 different rendered image sets. We report the mean global pairwise angular error over the whole set of 100 images over the training process in 
Table~\ref{tab:supervised_orientation}.

\subsection{Result Analysis}
We report results on three metrics: 1) mean global train absolute angular error; 2) median global train absolute angular error; 3) percentage of runs that converge with final pairwise angular error $<2^{\circ}$ after 10K steps, which is referred to as \textit{$2^{\circ}$ Acc}. Specifically, global relative angular error is calculated as the all-pair relative angular error for all pairs within the image set of 100. As see in Table~\ref{tab:supervised_orientation}, \textbf{Oracle-MRP} achieves comparable but larger mean and median pairwise angular error compared to \textbf{Oracle-Quaternion}, while both methods achieves the same \textit{$2^{\circ}$ Acc} of $87.5\%$. In summary, through this simple experiment, we find that MRP is able to achieve comparable but slightly worse train error for absolute orientation supervision compared to quaternions. Thus in the case of direct pose supervision, MRP may not be the best choice of rotation representation; using an open manifold such as in MRP is beneficial only in the case of relative pose supervision. 
\begin{table}[H]
    \centering
    \begin{tabular}{c|c c c}
    \hline
     & Mean & Median & \\
    Algorithm & Error (${}^{\circ}$) & Error (${}^{\circ}$) & $2^{\circ}$ Acc (\%)\\
    \hline
     \textbf{Oracle-Quaternion} & \textbf{1.58} & \textbf{1.56} &  \textbf{87.5} \\
     Oracle-MRP  & 1.81 & 1.86 & \textbf{87.5}\\
    \hline
    \end{tabular}
\caption{\textbf{Absolute Orientation Supervision for Image Based Rotation Estimation from Rendered YCB Drill Images using MRP vs Quaternions Parametrization.} Final mean, median angular train error ($\circ$) and convergence ($< 2^{\circ}$) percentage for image based rotation estimation from rendered YCB drill images with absolute orientation supervision, after 10K training steps over 8 sets of 100 rendered images.}
    \label{tab:supervised_orientation}
\end{table}

\section{Object Orientation Prediction Qualitative Visual Results}
\label{appendix:qualitative}

\begin{figure}[H]
 	\centering
     	\includegraphics[width=0.65\textwidth]{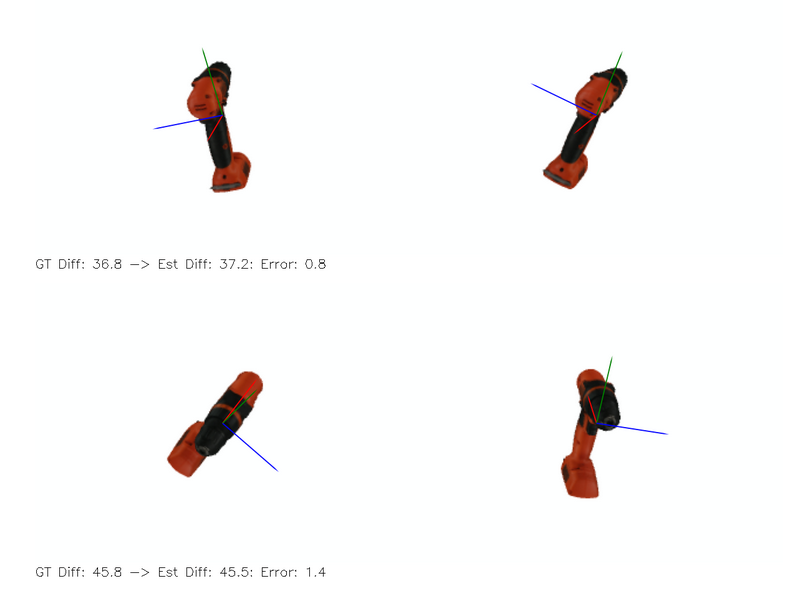}
 	\caption{\textbf{Qualitative Visual Examples for Object Orientation Estimation of MRP (Ours) on Rendered YCB Drill Images.} We show qualitative visual examples of predicted object 3D orientation from orientation prediction model trained via iterative MRP averaging with relative orientation supervision, the model is evaluated after training for 10K steps from neural net optimization experiment described Section 6.2 of the main paper. The predicted orientation is shown as coordinate frame (\textcolor{red}{x}, \textcolor{green}{y}, \textcolor{blue}{z}). On the bottom of each example, we show in text of the ground truth relative orientation angular difference (${}^{\circ}$) between the pair of images, and their predicted relative orientation angular difference (${}^{\circ}$) induced from the absolute object orientation predicted for each image. And finally we show the difference between the predicted relative angular difference and the ground truth relative angular difference as angular error (${}^{\circ}$).}
 
\label{fig:drill_visual}
 	 
\end{figure}

\begin{figure}[H]
 	\centering
     	\includegraphics[width=0.455\textwidth]{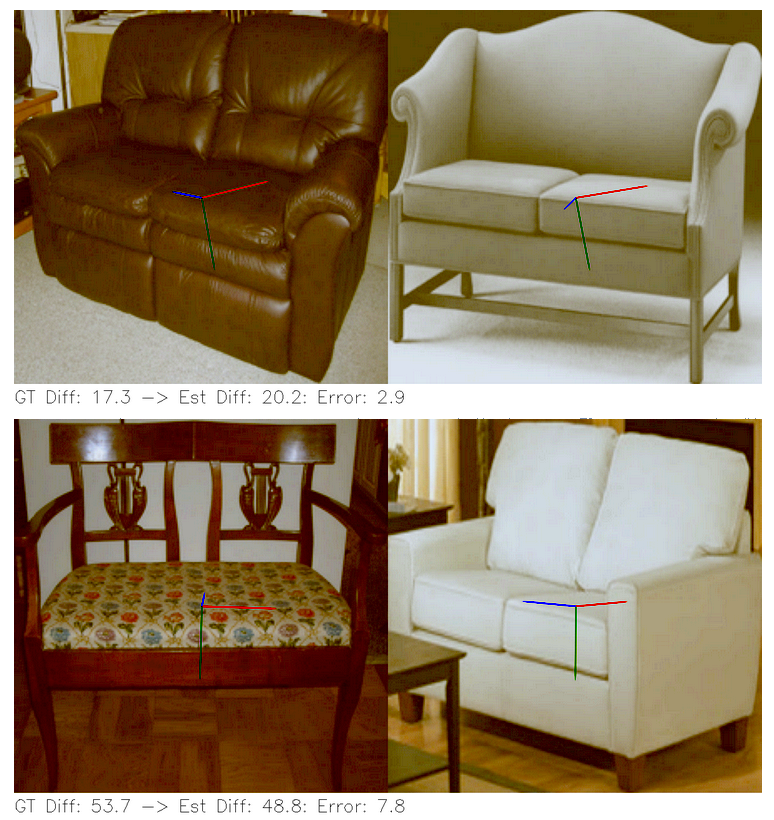}
        \hfill
     	\includegraphics[width=0.45\textwidth]{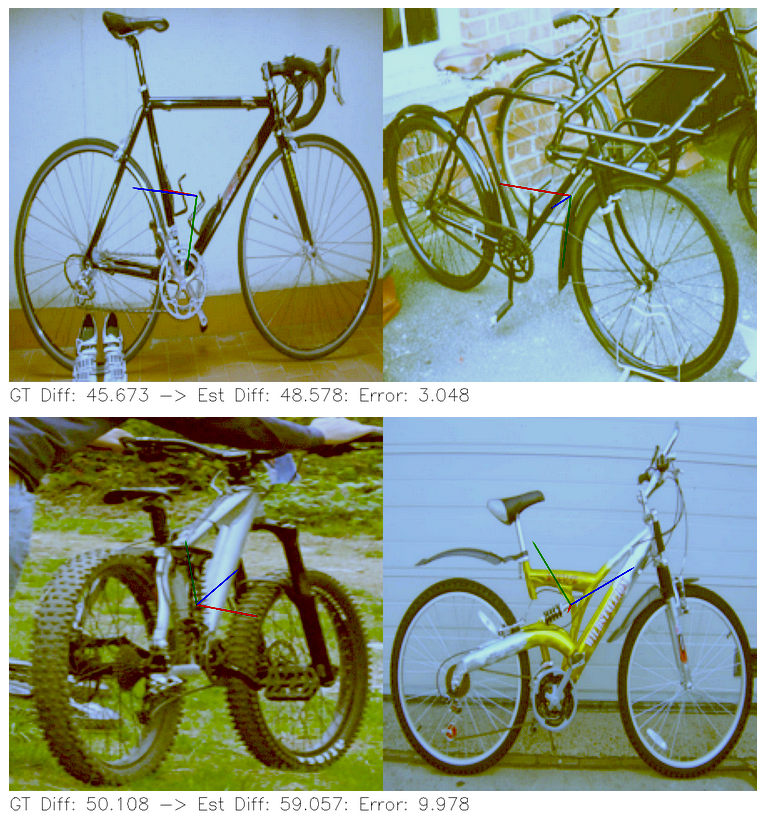}
 	\caption{\textbf{Qualitative Visual Examples for Object Orientation Estimation of MRP (Ours) on Unseen Pascal3D+ Images.}
 We show qualitative visual examples of predicted object 3D orientation from orientation prediction model trained via iterative MRP averaging with relative orientation supervision on \textit{Sofa} (left) and \textit{Bicycle} (right) images. The model is evaluated after training for 50K steps from 3D object rotation estimation on Pascal3D+ experiment as described Section~\ref{appendix:pascal_exp}. The predicted orientation is shown as coordinate frame (\textcolor{red}{x}, \textcolor{green}{y}, \textcolor{blue}{z}). On the bottom of each example, we show in text of the ground truth relative orientation angular difference~(${}^{\circ}$) between the pair of images, and their predicted relative orientation angular difference~(${}^{\circ}$) induced from the absolute object orientation predicted for each image. And finally we show the difference between the predicted relative angular difference and the ground truth relative angular difference as angular error (${}^{\circ}$).}
 
\label{fig:sofa_visual}
 	 
\end{figure}

We further show some qualitative visual illustrations of the object orientation prediction of trained model at convergence, trained using our iterative MRP averaging method via relative orientation supervision below. Examples from orientation estimation on the rendered YCB drill data as described in Section 6.2 in the main paper is shown in Figure~\ref{fig:drill_visual}. Examples from orientation estimation on unseen Pascal3D+ \textit{sofa} and \textit{bicycle} categories, as described in Section~\ref{appendix:pascal_exp_settings}, are shown in Figure~\ref{fig:sofa_visual}.
\putbib[root]
\end{bibunit}

\end{document}